\documentclass[10pt,twocolumn,letterpaper]{article}

\usepackage{cvpr}
\usepackage{times}
\usepackage{epsfig}
\usepackage{graphicx}
\usepackage{amsmath}
\usepackage{amssymb}
\usepackage{float}
\usepackage{color}
\usepackage{multirow}
\usepackage{array}
\newcolumntype{C}[1]{>{\centering\let\newline\\\arraybackslash\hspace{0pt}}m{#1}}
\usepackage{booktabs}
\usepackage{subfigure}
\usepackage[toc,page]{appendix}
\usepackage[pagebackref=true,breaklinks=true,letterpaper=true,colorlinks,bookmarks=false]{hyperref}

\cvprfinalcopy 

\def\cvprPaperID{} 

\ifcvprfinal\fi
\begin{document}

\title{\textit{PaSta}Net: Toward Human Activity Knowledge Engine}

\author{Yong-Lu Li,~Liang Xu,~Xinpeng Liu,~Xijie Huang,~Yue Xu,\\~Shiyi Wang,~Hao-Shu Fang,~Ze Ma,~Mingyang Chen,~Cewu Lu\thanks{Cewu Lu is the corresponding author, member of Qing Yuan Research Institute and MoE Key Lab of Artificial Intelligence, AI Institute, Shanghai Jiao Tong University, China.}\\
Shanghai Jiao Tong University\\
{\tt\small \{yonglu\_li, liangxu, otaku\_huang, silicxuyue, shiywang\}@sjtu.edu.cn,}\\
{\tt\small \{xinpengliu0907, fhaoshu\}@gmail.com,}
{\tt\small \{maze1234556, cmy\_123, lucewu\}@sjtu.edu.cn}
}
\maketitle
\thispagestyle{empty}

\begin{abstract}
Existing image-based activity understanding methods mainly adopt direct mapping, \ie from image to activity concepts, which may encounter performance bottleneck since the huge gap.
In light of this, we propose a new path: infer human part states first and then reason out the activities based on part-level semantics. 
Human Body Part States (\textbf{PaSta}) are fine-grained action semantic tokens, \eg $\langle hand, hold, something\rangle$, which can compose the activities and help us step toward human activity knowledge engine.
To fully utilize the power of \textit{PaSta}, we build a large-scale knowledge base \textbf{\textit{PaSta}Net}, which contains \textbf{7M+} \textit{PaSta} annotations. 
And two corresponding models are proposed: first, we design a model named Activity2Vec to extract \textit{PaSta} features, which aim to be general representations for various activities. Second, we use a \textit{PaSta}-based Reasoning method to infer activities. 
Promoted by \textit{PaSta}Net, our method achieves significant improvements, \eg 6.4 and 13.9 mAP on full and one-shot sets of HICO in supervised learning, and 3.2 and 4.2 mAP on V-COCO and images-based AVA in transfer learning.
Code and data are available at \url{http://hake-mvig.cn/}.
\end{abstract}
\vspace{-0.5cm}

\section{Introduction}
\begin{figure}[!ht]
	\begin{center}
		\includegraphics[width=0.45\textwidth]{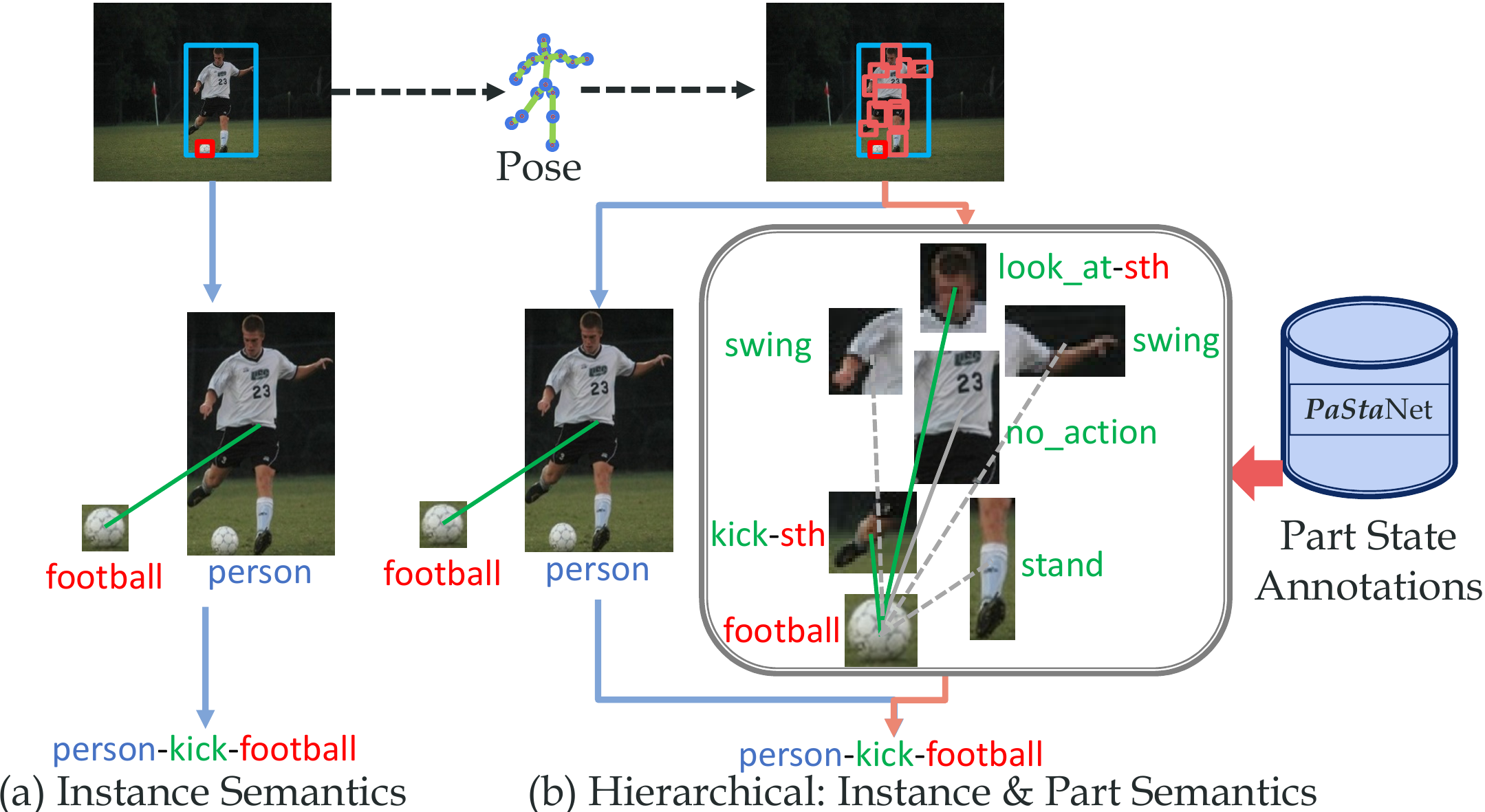}
	\end{center}
	\caption{Instance-level and hierarchical methods. Besides the instance-level path, we perform body part states recognition in part-level with the \textit{PaSta} annotations. 
	With the help of \textit{PaSta}, we can significantly boost the performance of activity understanding.}
	\label{Figure:paradigm_compare}
\vspace{-0.6cm}
\end{figure}
Understanding activity from images is crucial for building an intelligent system. Facilitated by deep learning, great advancements have been made in this field. 
Recent works~\cite{Delaitre2011Learning,Yao2012Action,Yang2010Recognizing,Maji2012Action} mainly address this high-level cognition task in one-stage, \ie from pixels to activity concept directly based on \textit{instance-level semantics} (Fig.~\ref{Figure:paradigm_compare}(a)). This strategy faces performance bottleneck on large-scale benchmarks~\cite{hicodet,AVA}. 
Understanding activities is difficult for reasons, \eg long-tail data distribution, complex visual patterns, \etc. 
Moreover, action understanding expects a \textit{knowledge engine} that can generally support activity related tasks. 
Thus, for data from another domain and unseen activities, much smaller effort is required for knowledge transfer and adaptation.
Additionally, for most cases, we find that only a few key human parts are relevant to the existing actions, the other parts usually carry very few useful clues.

Consider the example in Fig.~\ref{Figure:paradigm_compare}, we argue that perception in human \textit{part-level semantics} is a promising path but previously ignored.
Our core idea is that human instance actions are composed of \textit{fine-grained atomic body part states}. 
This lies in strong relationships with reductionism~\cite{Reductionism}.
Moreover, the part-level path can help us to pick up discriminative parts and disregard irrelevant ones. 
Therefore, encoding knowledge from human parts is a crucial step toward \emph{human activity knowledge engine}. 
The generic object part states~\cite{partstate} reveal that the semantic state of an object part is limited. For example, after exhaustively checking on $7M$ manually labeled body part state samples, we find that there are only about 12 states for ``head'' in daily life activities, such as ``listen to'', ``eat'', ``talk to'', ``inspect'', \etc.
Therefore, in this paper, we exhaustively collect and annotate the possible semantic meanings of human parts in activities to build a large-scale human part knowledge base \textbf{\textit{PaSta}Net} (\textit{PaSta} is the abbreviation of Body Part State). Now \textit{PaSta}Net includes \textbf{118 K+} images, \textbf{285 K+} persons, \textbf{250 K+} interacted objects, \textbf{724 K+} activities and \textbf{7 M+} human part states.
Extensive analysis verifies that \textit{PaSta}Net can cover most of the the part-level knowledge in general. Using learned \textit{PaSta} knowledge in \textit{transfer learning}, we can achieve 3.2, 4.2 and 3.2 improvements on V-COCO~\cite{vcoco}, images-based AVA~\cite{AVA} and HICO-DET~\cite{hicodet} (Sec.\ref{sec:transfer}).

Given \textit{PaSta}Net, we propose two powerful tools to promote the image-based activity understanding:
1) \textbf{Activity2Vec}: With \textit{PaSta}Net, we convert a human instance into a vector consisting of \textit{PaSta} representations. Activity2Vec extracts part-level semantic representation via \textit{PaSta} recognition and combines its language representation. Since \textit{PaSta} encodes common knowledge of activities, Activity2Vec works as a general feature extractor for both seen and unseen activities. 
2) \textbf{PaSta-R}:
A Part State Based Reasoning method (\textit{PaSta-R}) is further presented. We construct a \textit{Hierarchical Activity Graph} consisting of human instance and part semantic representations, and infer the activities by combining both instance and part level sub-graph states.

The advantages of our method are two-fold:
1) \textbf{Reusability and Transferability}: \textit{PaSta} are basic components of actions, their relationship can be in analogy with the amino acid and protein, letter and word, \etc. Hence, \textit{PaSta} are \textbf{reusable}, \eg, $\langle hand, hold, something\rangle$ is \textit{shared} by various actions like ``hold horse'' and ``eat apple''. 
Therefore, we get the capacity to describe and differentiate plenty of activities with a much smaller set of \textit{PaSta}, \ie one-time labeling and transferability. For few-shot learning, reusability can greatly alleviate its learning difficulty. Thus our approach shows significant improvements, \eg we boost 13.9 mAP on one-shot sets of HICO~\cite{hico}.
2) \textbf{Interpretability}: we obtain not only more powerful activity representations, but also better interpretation. When the model predicts what a person is doing, we can easily know the reasons: what the body parts are doing.

In conclusion, we believe \textit{PaSta}Net will function as a human activity knowledge engine. Our main contributions are:
1) We construct \textit{PaSta}Net, the first large-scale activity knowledge base with fine-grained \textit{PaSta} annotations.
2) We propose a novel method to extract part-level activity representation named Activity2Vec and a \textit{PaSta}-based reasoning method.
3) In supervised and transfer learning, our method achieves significant improvements on large-scale activity benchmarks, \eg 6.4 (16\%), 5.6 (33\%) mAP improvements on HICO~\cite{hico} and HICO-DET~\cite{hicodet} respectively. 

\section{Related Works} 
\noindent{\bf Activity Understanding.} Benefited by deep learning and large-scale datasets, image-based~\cite{hico,vcoco,openimages,MPII} or video-based~\cite{AVA,pang2019deep,pang2020deep,Kinetics,UCF101,shao2018find,JhuangICCV2013} activity understanding has achieved huge improvements recently.
Human activities have a hierarchical structure and include diverse verbs, so it is hard to define an explicit organization for their categories. 
Existing datasets~\cite{AVA,hico,activitynet,vcoco} often have a large difference in definition, thus transferring knowledge from one dataset to another is ineffective. 
Meanwhile, plenty of works have been proposed to address the activity understanding~\cite{Delaitre2011Learning, Gkioxari_2015_ICCV, Du_2015_CVPR, NIPS2014_5353, Feichtenhofer_2016_CVPR, 6165309, Tran_2015_ICCV, Sun_2015_ICCV}. 
There are holistic body-level approaches~\cite{Thurau2008Pose, Delaitre2011Learning}, body part-based methods~\cite{Gkioxari_2015_ICCV}, and skeleton-based methods~\cite{Vemulapalli_2014_CVPR, Du_2015_CVPR}, \etc. 
But compared with other tasks such as object detection~\cite{faster} or pose estimation~\cite{fang2017rmpe}, its performance is still limited.

\noindent{\bf Human-Object Interaction.} Human-Object Interaction (HOI)~\cite{hico,hicodet} occupies the most of daily human activities.
In terms of tasks, some works focus on image-based HOI recognition~\cite{hico}. Furthermore, instance-based HOI detection~\cite{hicodet,vcoco} needs to detect accurate positions of the humans and objects and classify interaction simultaneously. 
In terms of the information utilization, some works utilized holistic human body and pose~\cite{Thurau2008Pose,Yao2012Action,Yang2010Recognizing,Maji2012Action,choutas2018potion}, and global context is also proved to be effective~\cite{Hu2014Recognising,Yao2010Modeling,Yao2011Human,Delaitre2011Learning}. 
According to the learning paradigm, earlier works were often based on hand-crafted features~\cite{Delaitre2011Learning,Hu2014Recognising}. Benefited from large scale HOI datasets, recent approaches~\cite{Gkioxari_2015_ICCV,Fang2018Pairwise,Gkioxari2014R,Gkioxari2017Detecting,Mallya2016Learning,qi2018learning,gao2018ican,interactiveness} started to use deep neural networks to extract features and achieved great improvements.

\noindent{\bf Body Part based Methods.} Besides the instance pattern, some approaches studied to utilize part pattern~\cite{Yao2010Modeling,Fang2018Pairwise,Gkioxari_2015_ICCV,Du_2015_CVPR,Maji2012Action,zhao2017single}.
Gkioxari~\etal~\cite{Gkioxari_2015_ICCV} detects both the instance and parts and input them all into a classifier.
Fang~\etal~\cite{Fang2018Pairwise} defines part pairs and encodes pair features to improve HOI recognition.
Yao~\etal~\cite{Yao2010Modeling} builds a graphical model and embed parts appearance as nodes, and use them with object feature and pose to predict the HOIs.
Previous work mainly utilized the part \textit{appearance} and \textit{location}, but few studies tried to divide the instance actions into discrete part-level semantic tokens, and refer them as the basic components of activity concepts. 
In comparison, we aim at building human part semantics as \textbf{reusable} and \textbf{transferable} knowledge.

\noindent{\bf Part States.} Part state is proposed in~\cite{partstate}. By tokenizing the semantic space as a discrete set of part states, \cite{partstate} constructs a sort of basic descriptors based on segmentation~\cite{he2017mask,fang2019instaboost,xu2018srda}. 
To exploit this cue, we divide the human body into natural parts and utilize their states as discretized part semantics to represent activities. 
In this paper, we focus on the part states of humans instead of daily objects.

\section{Constructing \textit{PaSta}Net} 
\label{sec:hake-construction}
In this section, we introduce the construction of \textit{PaSta}Net. \textit{PaSta}Net seeks to explore the common knowledge of human \textit{PaSta} as atomic elements to infer activities. 

\noindent{\bf \textit{PaSta} Definition.} We decompose human body into \textit{ten} parts, namely \textit{head, two upper arms, two hands, hip, two thighs, two feet}. 
Part states (\textit{PaSta}) will be assigned to these parts. Each \textit{PaSta} represents a description of the target part. For example, the \textit{PaSta} of ``hand'' can be ``hold something'' or ``push something'', the \textit{PaSta} of ``head'' can be ``watch something'', ``eat something''. 
After exhaustively reviewing collected 200K+ images, we found the descriptions of any human parts can be concluded into limited categories. That is, the \textit{PaSta} category number of each part is limited. 
Especially, a person may have more than one action simultaneously, thus each part can have multiple \textit{PaSta}, too.   

\noindent{\bf Data Collection.}
For generality, we collect human-centric activity images by crowdsourcing (30K images paired with rough activity label) as well as existing well-designed datasets~\cite{hico,hicodet,vcoco,openimages,hcvrd,pic} (185K images), which are structured around a rich semantic ontology, diversity, and variability of activities. All their annotated persons and objects are extracted for our construction. 
Finally, we collect more than 200K images of diverse activity categories. 

\noindent{\bf Activity Labeling.} 
Activity categories of \textit{PaSta}Net are chosen according to the most common  \textit{human daily activities, interactions with object and person}. 
Referred to the hierarchical activity structure~\cite{activitynet}, common activities in existing datasets~\cite{hico,vcoco,hcvrd,openimages,AVA,activitynet,MPII,pic} and crowdsourcing labels, we select 156 activities including human-object interactions and body motions from 118K images.
According to them, we first clean and reorganize the annotated human and objects from existing datasets and crowdsourcing.
Then, we annotate the active persons and the interacted objects in the rest of the images. 
Thus, \textit{PaSta}Net includes all active human and object bounding boxes of 156 activities.

\noindent{\bf Body Part Box.} 
To locate the human parts, we use pose estimation~\cite{fang2017rmpe} to obtain the joints of all annotated persons. Then we generate \textit{ten} body part boxes following~\cite{Fang2018Pairwise}.
Estimation errors are addressed manually to ensure high-quality annotation. 
Each part box is centered with a joint, and the box size is pre-defined by scaling the distance between the joints of the neck and pelvis. 
A joint with confidence higher than 0.7 will be seen as visible. When not all joints can be detected, we use \textit{body knowledge-based rules}. 
That is, if the neck or pelvis is invisible, we configure the part boxes according to other visible joint groups (head, main body, arms, legs), \eg, if only the upper body is visible, we set the size of the hand box to twice the pupil distance.

\begin{figure}[!ht]
	\begin{center}
		\includegraphics[width=0.45\textwidth]{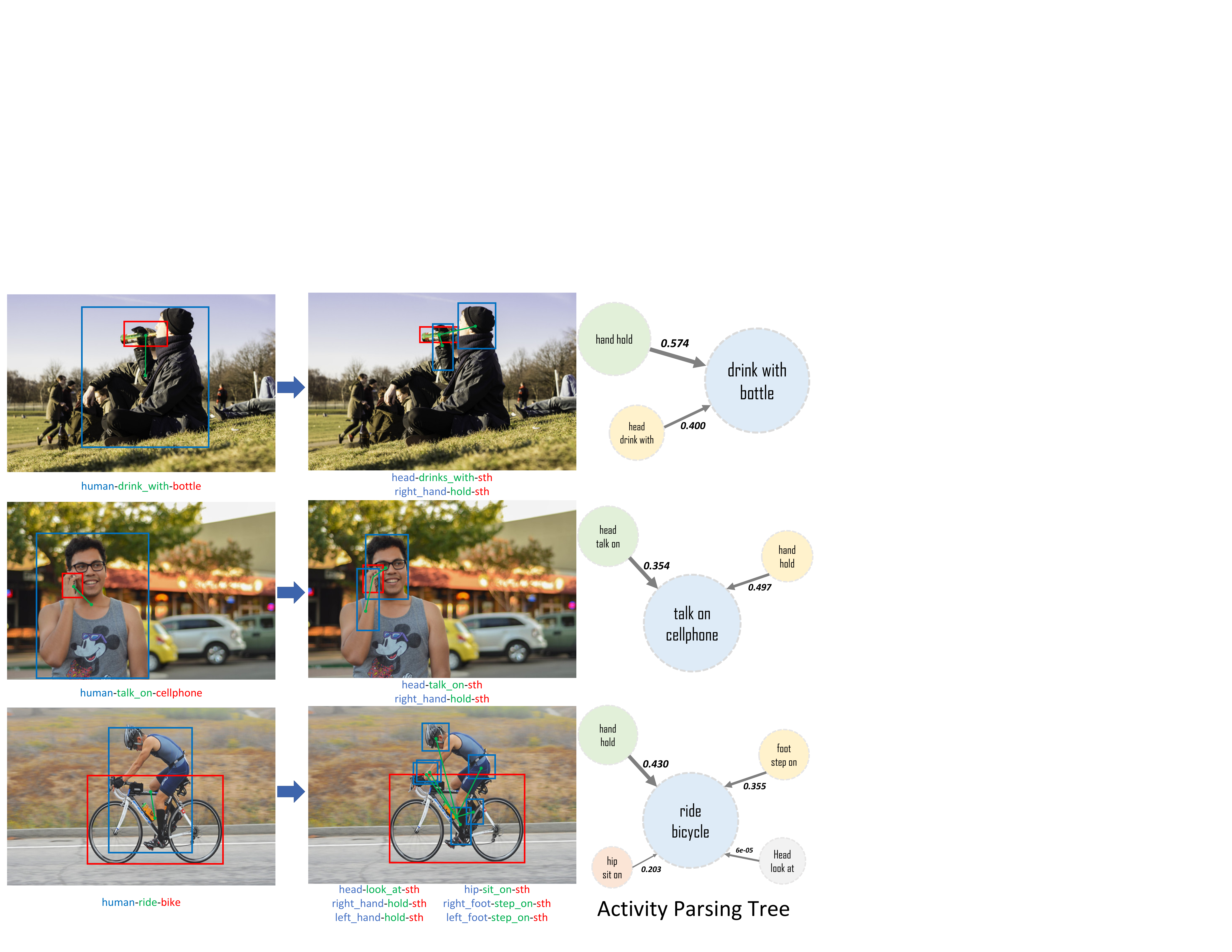}
	\end{center}
	\caption{\textit{PaSta} annotation. Based on instance activity labels, we add fine-grained body part boxes and corresponding part states \textit{PaSta} labels. In \textit{PaSta}, we use ``something''~\cite{partstate} to indicate the interacted object for generalization. The edge in \textit{Activity Parsing Tree} indicates the statistical co-occurrence.}
	\label{Figure:pasta-example}
\vspace{-0.7cm}
\end{figure}

\noindent{\bf \textit{PaSta} Annotation.} 
We carry out the annotation by crowdsourcing and receive 224,159 annotation uploads. The process is as follows:
1) First, we choose the \textit{PaSta} categories considering the \textit{generalization}.
Based on the verbs of 156 activities, we choose 200 verbs from WordNet~\cite{miller1995wordnet} as the \textit{PaSta} candidates, \eg, ``hold'', ``pick'' for hands, ``eat'', ``talk to'' for head, \etc. If a part does not have any active states, we depict it as ``no\_action''.
2) Second, to find the most common \textit{PaSta} that can work as the \textit{transferable activity knowledge}, we invite \textbf{150} annotators from different backgrounds to annotate 10K images of 156 activities with \textit{PaSta} candidates (Fig.~\ref{Figure:pasta-example}). 
For example, given an activity ``ride bicycle'', they may describe it as $\langle hip, sit\_on, something\rangle$, $\langle hand, hold, something\rangle$, $\langle foot, step\_on, something\rangle$, \etc.
3) Based on their annotations, we use the Normalized Point-wise Mutual Information (NPMI)~\cite{church1990word} to calculate the co-occurrence between activities and \textit{PaSta} candidates. Finally, we choose \textbf{76} candidates with the highest NPMI values as the final \textit{PaSta}. 
4) Using the annotations of 10K images as seeds, we automatically generate the initial \textit{PaSta} labels for all of the rest images. Thus the other 210 annotators only need to revise the annotations.
5) Considering that a person may have multiple actions, for \textit{each} action, we annotate its corresponding ten \textit{PaSta} respectively. 
Then we combine all sets of \textit{PaSta} from all actions. Thus, a part can also have multiple states, \eg, in ``eating while talking'', the head has \textit{PaSta} $\langle head, eat, something\rangle$, $\langle head, talk\_to, something\rangle$ and $\langle head, look\_at, something\rangle$ simultaneously.
6) To ensure quality, each image will be annotated twice and checked by automatic procedures and supervisors. We cluster all labels and discard the outliers to obtain robust agreements.

\noindent{\bf Activity Parsing Tree.} 
To illustrate the relationships between \textit{PaSta} and activities, we use their statistical correlations to construct a graph (Fig.~\ref{Figure:pasta-example}): activities are root nodes, \textit{PaSta} are son nodes and edges are co-occurrence.

Finally, \textit{PaSta}Net includes \textbf{118K+} images, \textbf{285K+} persons, \textbf{250K+} interacted objects, \textbf{724K+} instance activities and \textbf{7M+} \textit{PaSta}.
Referred to well-designed datasets~\cite{AVA,activitynet,hico} and WordNet~\cite{miller1995wordnet}, \textit{PaSta} can cover most part situations with good \textbf{generalization}.
To verify that \textit{PaSta} have encoded common part-level activity knowledge and can adapt to various activities, we adopt two experiments:

\noindent{\bf Coverage Experiment.}
To verify that \textit{PaSta} can cover most of the activities, we collect other 50K images out of \textit{PaSta}Net. Those images contain diverse activities and many of them are unseen in \textit{PaSta}Net. 
Another 100 volunteers from different backgrounds are invited to find human parts that can not be well described by our \textit{PaSta} set. We found that only $2.3 \%$ cases cannot find appropriate descriptions. This verifies that \textit{PaSta}Net is general to activities. 

\noindent{\bf Recognition Experiment.} 
First, we find that \textit{PaSta} can be well \textbf{learned}. 
A shallow model trained with a part of \textit{PaSta}Net can easily achieve about \textbf{55} mAP on \textit{PaSta} recognition.
Meanwhile, a deeper model can only achieve about 40 mAP on activity recognition with the same data and metric (Sec.~\ref{sec:experiment-hico}).
Second, we argue that \textit{PaSta} can be well \textbf{transferred}.
To verify this, we conduct transfer learning experiments (Sec.~\ref{sec:transfer}), \ie first trains a model to learn the knowledge from \textit{PaSta}Net, then use it to infer the activities of unseen datasets, even unseen activities. Results show that \textit{PaSta} can be well transferred and boost the performance (4.2 mAP on image-based AVA). 
Thus it can be considered as the general part-level activity knowledge.
\begin{figure*}[!ht]
	\begin{center}
		\includegraphics[width=0.9\textwidth]{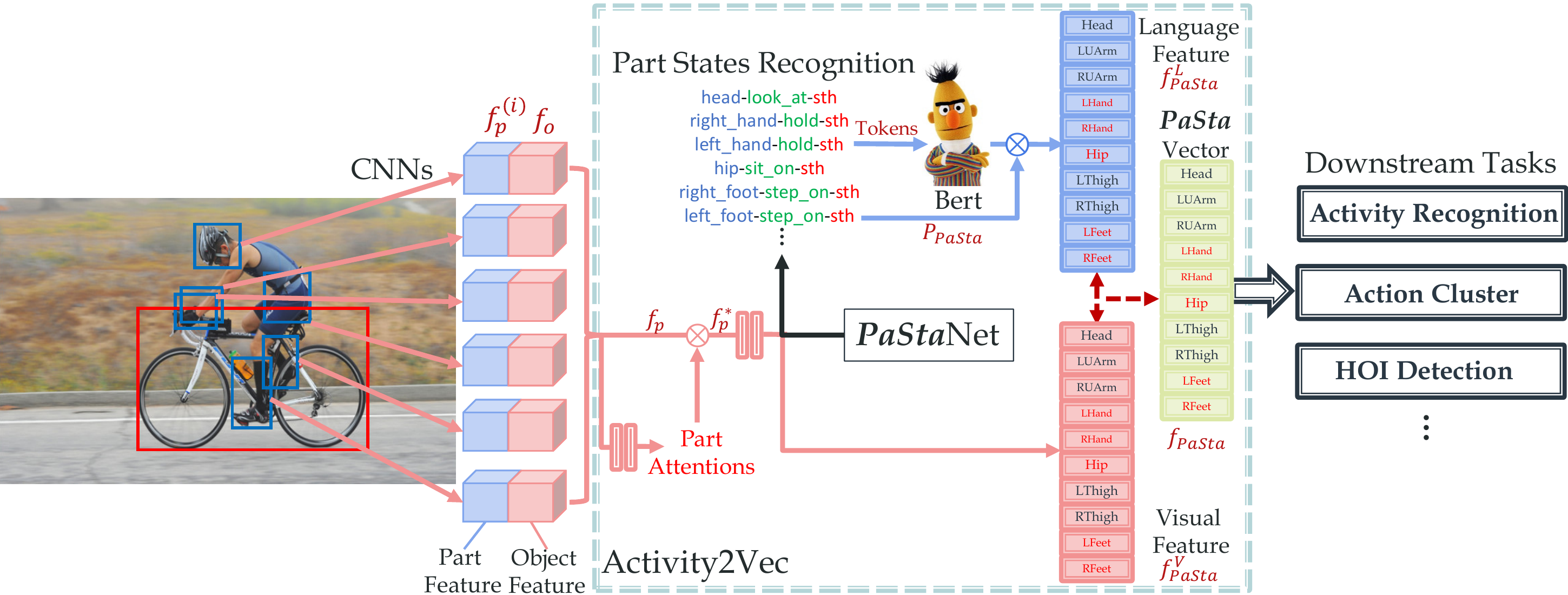}
	\end{center}
	\caption{The overview of Part States (\textit{PaSta}) recognition and Activity2Vec.}
	\label{Figure:overview}
\vspace{-0.3cm}
\end{figure*}

\section{Activity Representation by \textit{PaSta}Net}
In this section, we discuss the activity representation by \textit{PaSta}Net. 

\noindent{\bf Conventional Paradigm} Given an image $I$, conventional methods mainly use a direct mapping (Fig.~\ref{Figure:paradigm_compare}(a)):
\begin{eqnarray}
\label{eq:previous}
    \mathcal{S}_{inst}=\mathcal{F}_{inst}(I, b_h, \mathcal{B}_o)
\end{eqnarray}
to infer the action score $\mathcal{S}_{inst}$ with instance-level semantic representations $f_{inst}$. $b_h$ is the human box and $\mathcal{B}_o=\{b_o^i\}^{m}_{i=1}$ are the $m$ interacted object boxes of this person.

\noindent{\bf \textit{PaSta}Net Paradigm.} We propose a novel paradigm to utilize general part knowledge: 
1) \textit{PaSta} recognition and feature extraction for a person and an interacted object $b_o$:
\begin{eqnarray}
\label{eq:a2v}
    f_{PaSta} = \mathcal{R}_{A2V}(I, \mathcal{B}_p, b_o),
\end{eqnarray}
where $\mathcal{B}_p=\{b^{(i)}_p\}^{10}_{i}$ are part boxes generated from the pose estimation~\cite{fang2017rmpe} automatically following \cite{Fang2018Pairwise} (head, upper arms, hands, hip, thighs, feet).
$\mathcal{R}_{A2V}(\cdot)$ indicates the Activity2Vec, which extracts ten \textit{PaSta} representations $f_{PaSta}=\{f^{(i)}_{PaSta}\}^{10}_{i=1}$.
2) \textit{PaSta}-based Reasoning (\textit{PaSta-R}), \ie, from \textit{PaSta} to activity semantics:
\begin{eqnarray}
\label{eq:reasoning}
    \mathcal{S}_{part} = \mathcal{F}_{PaSta-R}(f_{PaSta}, f_o),
\end{eqnarray}
where $\mathcal{F}_{PaSta-R}(\cdot)$ indicates the \textit{PaSta-R}, $f_o$ is the object feature. $\mathcal{S}_{part}$ is the action score of the part-level path.
If the person does not interact with any objects, we use the ROI pooling feature of the whole image as $f_o$. 
For multiple object case, \ie, a person interacts with several objects, 
we process each human-object pair $(f_{PaSta}, f_o^{(i)})$ respectively and generate its Activity2Vec embedding.

Following, we introduce the \textit{PaSta} recognition in Sec.~\ref{sec:PS}. 
Then, we discuss how to map human instance to semantic vector via Activity2Vec in Sec.~\ref{sec:a2v}. We believe it can be a general activity representation extractor. In Sec.~\ref{sec:pasta-r}, a hierarchical activity graph is proposed to largely advance activity related tasks by leveraging \textit{PaSta}Net.  

\subsection{Part State Recognition} 
\label{sec:PS}
With the object and body part boxes $b_o, \mathcal{B}_p$, we operate the \textit{PaSta} recognition as shown in Fig.~\ref{Figure:overview}.
In detail, a COCO~\cite{coco} pre-trained Faster R-CNN~\cite{faster} is used as the feature extractor.
For each part, we concatenate the part feature $f^{(i)}_p$ from $b^{(i)}_p$ and object features $f_o$ from $b_o$ as inputs. For body only motion, we input the whole image feature $f_c$ as $f_o$.
All features will be first input to a \textbf{Part Relevance Predictor}.
Part relevance represents how important a body part is to the action. 
For example, feet usually have weak correlations with ``drink with cup''. And in ``eat apple'', only hands and head are essential.
These relevance/attention labels can be converted from \textit{PaSta} labels directly, \ie the attention label will be \textit{one}, unless its \textit{PaSta} label is ``no\_action'', which means this part contributes nothing to the action inference.
With the part attention labels as supervision, we use part relevance predictor consisting of FC layers and Sigmoids to infer the attentions $\{a_{i}\}^{10}_{i=1}$ of each part.
Formally, for a person and an interacted object:
\begin{eqnarray}
\label{eq:part-loss}
    a_{i} = \mathcal{P}_{pa}(f^{(i)}_p, f_o),
\end{eqnarray}
where $\mathcal{P}_{pa}(\cdot)$ is the part attention predictor.
We compute cross-entropy loss $\mathcal{L}^{(i)}_{att}$ for each part and multiply $f^{(i)}_p$ with its scalar attention, \ie $f^{(i)*}_p=f^{(i)}_p \times a_{i}$. 

Second, we operate the \textit{PaSta} recognition. 
For each part, we concatenate the re-weighted $f^{(i)*}_p$ with $f_o$, and input them into a max pooling layer and two subsequent 512 sized FC layers, thus obtain the \textit{PaSta} score $\mathcal{S}^{(i)}_{PaSta}$ for the $i$-th part.
Because a part can have multiple states, \eg head performs ``eat'' and ``watch'' simultaneously.
Hence we use multiple Sigmoids to do this multi-label classification.
With \textit{PaSta} labels, we construct cross-entropy loss $\mathcal{L}^{(i)}_{PaSta}$.
The total loss of \textit{PaSta} recognition is:
\begin{eqnarray}
\label{eq:part-loss}
    \mathcal{L}_{PaSta} = \sum^{10}_{i} (\mathcal{L}^{(i)}_{PaSta} + \mathcal{L}^{(i)}_{att}).
\end{eqnarray}

\subsection{Activity2Vec}
\label{sec:a2v}
In Sec.~\ref{sec:hake-construction}, we define the \textit{PaSta} according to the most common activities. That is, choosing the part-level verbs which are \textit{most often used to compose and describe the activities} by a large number of annotators. Therefore \textit{PaSta} can be seen as the fundamental components of instance activities.
Meanwhile, \textit{PaSta} recognition can be well learned. Thus, we can operate \textit{PaSta} recognition on \textit{PaSta}Net to learn the powerful \textit{PaSta} representations, which have good transferability. They can be used to reason out the instance actions in both supervised and transfer learning.
Under such circumstance, \textit{PaSta}Net works like the ImageNet~\cite{imagenet}. And \textit{PaSta}Net pre-trained Activity2Vec functions as a knowledge engine and transfers the knowledge to other tasks.

\noindent{\bf Visual \textit{PaSta} feature.}
First, we extract visual \textit{PaSta} representations from \textit{PaSta} recognition.
Specifically, we extract the feature from the last FC layer in \textit{PaSta} classifier as the visual \textit{PaSta} representation $f^{V(i)}_{PaSta} \in \mathbb{R}^{512}$.

\noindent{\bf Language \textit{PaSta} feature.}
Our goal is to bridge the gap between \textit{PaSta} and activity semantics. 
Language priors are useful in visual concept understanding~\cite{Lu2016Visual,vinyals2015show}.
Thus the combination of visual and language knowledge is a good choice for establishing this mapping.
To further enhance the representation ability, we utilize the uncased BERT-Base pre-trained model~\cite{devlin2018bert} as the language representation extractor. 
Bert~\cite{devlin2018bert} is a language understanding model that considers the context of words and uses a deep bidirectional transformer to extract contextual representations.
It is trained with large-scale corpus databases such as Wikipedia, hence the generated embedding contains helpful implicit semantic knowledge about the activity and \textit{PaSta}.
For example, the description of the entry ``basketball'' in Wikipedia: ``drag one's \textit{foot} without \textit{dribbling} the ball, to \textit{carry} it, or to \textit{hold} the ball with both \textit{hands}...\textit{placing} his \textit{hand} on the bottom of the ball;..known as \textit{carrying} the ball''.

In specific, for the $i$-th body part with $n$ \textit{PaSta}, 
we divide each \textit{PaSta} into tokens $\{t^{(i,k)}_p, t^{(i,k)}_v, t^{(i,k)}_o\}^{n}_{k=1}$, \eg, $\langle part, verb, object\rangle$. 
The $\langle object \rangle$ comes from object detection.
Each \textit{PaSta} will be converted to a $f^{(i,k)}_{Bert} \in \mathbb{R}^{2304}$ (concatenating three 768 sized vectors of part, verb, object), \ie $f^{(i,k)}_{Bert}=\mathcal{R}_{Bert}(t^{(i,k)}_p, t^{(i,k)}_v, t^{(i,k)}_o)$.  
$\{f^{(i,k)}_{Bert}\}^{n}_{k=1}$ will be concatenated as the $f^{(i)}_{Bert} \in \mathbb{R}^{2304*n}$ for the $i$-th part.
Second, we multiply $f^{(i)}_{Bert}$ with predicted \textit{PaSta} probabilities $P^{(i)}_{PaSta}$, \ie $f^{L(i)}_{PaSta}=f^{(i)}_{Bert} \times P^{(i)}_{PaSta}$, where $P^{(i)}_{PaSta}=Sigmoid(\mathcal{S}^{(i)}_{PaSta}) \in \mathbb{R}^{n}$, $\mathcal{S}_{PaSta}^{(i)}$ denotes the \textit{PaSta} score of the $i$-th part, $P_{PaSta}=\{P^{(i)}_{PaSta}\}^{10}_{i=1}$. 
This means a more possible \textit{PaSta} will get larger attention.
$f^{L(i)}_{PaSta} \in \mathbb{R}^{2304*n}$ is the final language \textit{PaSta} feature of the $i$-th part.
We use the pre-converted and frozen $f^{(i,k)}_{Bert}$ in the whole process.
Additionally, we also try to rewrite each \textit{PaSta} into a sentence and convert it into a fixed-size vector as $f^{(i,k)}_{Bert}$ and the performance is slightly better (Sec.~\ref{sec:ablation}).

\noindent{\bf \textit{PaSta} Representation.}
At last, we pool and resize the $f^{L(i)}_{PaSta}$, and concatenate it with its corresponding visual \textit{PaSta} feature $f^{V(i)}_{PaSta}$. 
Then we obtain the \textit{PaSta} representation $f^{(i)}_{PaSta} \in \mathbb{R}^{m}$ for each body part (\eg $m=4096$).
This process is indicated as \textbf{Activity2Vec} (Fig.~\ref{Figure:overview}).
The output $f_{PaSta}=\{{f^{(i)}_{PaSta}}\}^{10}_{i=1}$ is the part-level activity representation and can be used for various downstream tasks, \eg activity detection, captioning, \etc.
From the experiments, we can find that Activity2Vec has a powerful representational capacity and can significantly improve the performance of activity related tasks. 
It works like a knowledge transformer with the fundamental \textit{PaSta} to compose various activities.

\subsection{\textit{PaSta}-based Activity Reasoning}
\label{sec:pasta-r}
With part-level $f_{PaSta}$, we construct a Hierarchical Activity Graph (HAG) to model the activities. Then we can extract the graph state to reason out the activities.

\begin{figure}[!ht]
	\begin{center}
		\includegraphics[width=0.49\textwidth]{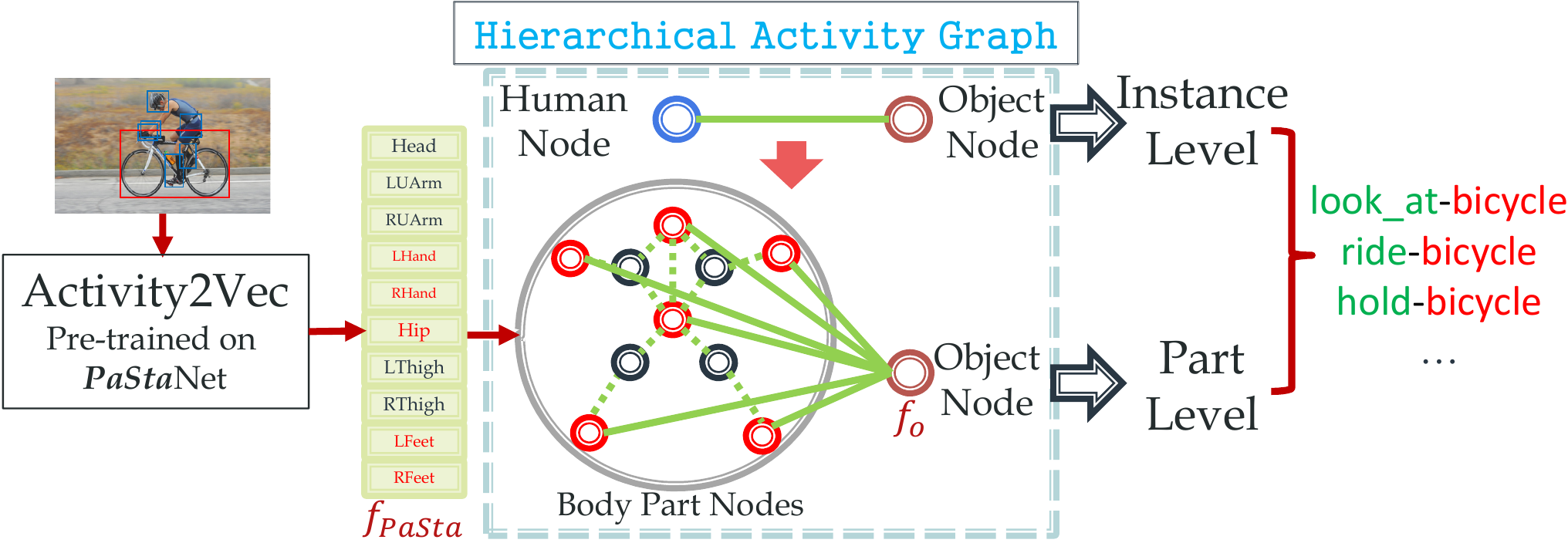}
	\end{center}
	\caption{From \textit{PaSta} to activities on hierarchical activity graph.}
	\label{Figure:reasoning}
\vspace{-0.5cm}
\end{figure}

\noindent{\bf Hierarchical Activity Graph.}
\label{sec:hag}
Hierarchical activity graph $\mathcal{G} = (\mathcal{V}, \mathcal{E})$ is depicted in Fig.~\ref{Figure:reasoning}.
For human-object interactions, $\mathcal{V} = \{\mathcal{V}_p, \mathcal{V}_o\}$. 
For body only motions, $\mathcal{V} = \mathcal{V}_p$.
In instance level, a person is a node with instance representation from previous instance-level methods~\cite{gao2018ican,interactiveness,AVA} as a node feature.
Object node $v_o \in \mathcal{V}_o$ and has $f_o$ as node feature. 
In part level, each body part can be seen as a node $v^{i}_p \in \mathcal{V}_p$ with \textit{PaSta} representation $f^{i}_{PaSta}$ as node feature. 
Edge between body parts and object is $e_{po} = (v^{i}_p, v_o) \in \mathcal{V}_p \times \mathcal{V}_o$, and edge within parts is $e^{ij}_{pp} = (v^{i}_p, v^{j}_p) \in \mathcal{V}_p \times \mathcal{V}_p$.

Our goal is to parse HAG and reason out the graph state, \ie activities.
In part-level, we use \textit{PaSta}-based Activity Reasoning (\textit{PaSta-R}) to infer the activities.
That is, with the \textit{PaSta} representation from Activity2Vec, we use $\mathcal{S}_{part} = \mathcal{F}_{PaSta-R}(f_{PaSta}, f_o)$ (Eq.~\ref{eq:reasoning}) to infer the activity scores $\mathcal{S}_{part}$.
For body motion only activities \eg ``dance'', Eq.~\ref{eq:reasoning} is $\mathcal{S}_{part} = \mathcal{F}_{PaSta-R}(f_{PaSta}, f_c)$, $f_c$ is the feature of image.
We adopt different implementations of $\mathcal{F}_{PaSta-R}(\cdot)$. 

\noindent{\bf Linear Combination.} The simplest implementation is to directly combine the part node features linearly. We concatenate the output of Activity2Vec $f_{PaSta}$ with $f_o$ and input them to a FC layer with Sigmoids.

\noindent{\bf MLP.} We can also operate nonlinear transformation on Activity2Vec output. We use two 1024 sized FC layers and an action category sized FC with Sigmoids.

\noindent{\bf Graph Convolution Network.} With part-level graph, we use Graph Convolution Network (GCN)~\cite{gcn} to extract the global graph feature and use an MLP subsequently.

\noindent{\bf Sequential Model.} When watching an image in this way: watch body part and object patches with language description one by one, human can easily guess the actions. 
Inspired by this, we adopt an LSTM~\cite{lstm} to take the part node features ${f^{(i)}_{PaSta}}$ gradually, and use the output of the last time step to classify actions. 
We adopt two input orders: random and fixed (from head to foot), and fixed order is better.

\noindent{\bf Tree-Structured Passing.} Human body has a natural hierarchy. Thus we use a tree-structured graph passing. 
Specifically, we first combine the hand and upper arm nodes into an ``arm'' node, its feature is obtained by concatenating the features of three son nodes and passed a 512 sized FC layer. 
Similarly, we combine the foot and thigh nodes to an ``leg'' node. Head, arms, legs and feet nodes together form the second level. 
The third level contains the ``upper body`` (head, arms) and ``lower-body'' (hip, legs). 
Finally, the body node is generated. We input it and the object node into an MLP.
\begin{figure}[!ht]
	\begin{center}
		\includegraphics[width=0.45\textwidth]{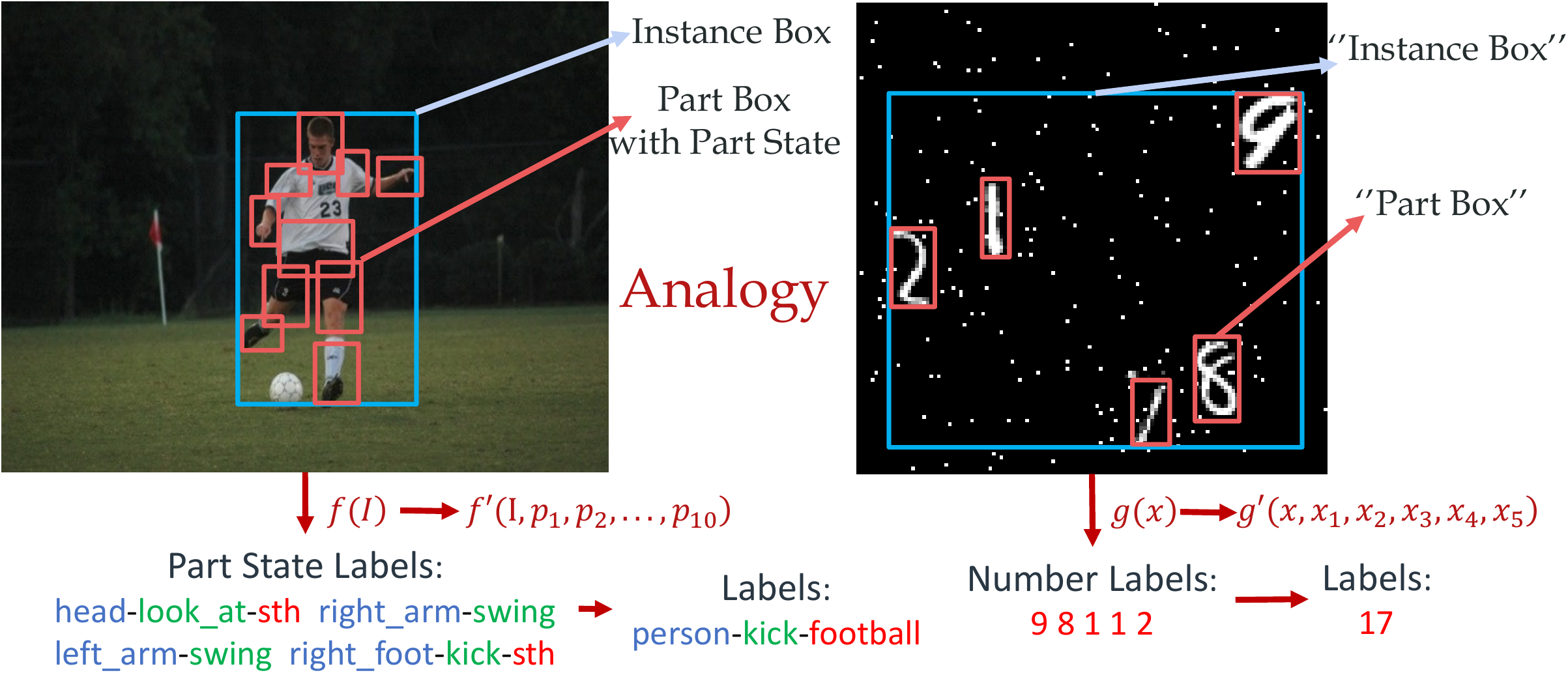}
	\end{center}
	\caption{An analogy to activity recognition.}
	\label{Figure:analogy}
\vspace{-0.5cm}
\end{figure}

The instance-level graph inference can be operated by instance-based methods~\cite{Fang2018Pairwise,gao2018ican,interactiveness,AVA} using Eq.~\ref{eq:previous}: $\mathcal{S}_{inst}=\mathcal{F}_{inst}(I, b_h, \mathcal{B}_o)$.
To get the final result upon the whole graph, we can use either early or late fusion.
In early fusion, we concatenate $f_{inst}$ with $f_{PaSta},f_o$ and input them to \textit{PaSta-R}.
In late fusion, we fuse the predictions of two levels, \ie $\mathcal{S} = \mathcal{S}_{inst} + \mathcal{S}_{part}$.
In our test, late fusion outperforms early fusion in most cases. 
If not specified, we use late fusion in Sec.~\ref{sec:experiments}.
We use $\mathcal{L}^{inst}_{cls}$ and $\mathcal{L}^{PaSta}_{cls}$ to indicate the cross-entropy losses of two levels. 
The total loss is:
\begin{eqnarray}
\label{eq:total-loss}
    \mathcal{L}_{total} = \mathcal{L}_{PaSta} + \mathcal{L}^{PaSta}_{cls} + \mathcal{L}^{inst}_{cls}.
\end{eqnarray}

\section{Experiments}
\label{sec:experiments}
\subsection{An analogy: MNIST-Action} 
We design a simplified experiment to give an intuition (Fig.~\ref{Figure:analogy}).
We randomly sample MNIST digits from 0 to 9 ($28 \times 28 \times 1$) and generate $128 \times 128 \times 1$ images consists of 3 to 5 digits.
Each image is given a label to indicate \textit{the sum of the two largest numbers} within it (0 to 18).
We assume that ``\textit{PaSta}-Activity'' resembles the ``Digits-Sum''. 
Body parts can be seen as digits, thus human is the union box of all digits.
To imitate the complex body movements, digits are randomly distributed, and Gaussian noise is added to the images.
For comparison, we adopt two simple networks. 
For instance-level model, we input the ROI pooling feature of the digit union box into an MLP.
For hierarchical model, we operate single-digit recognition, then concatenate the union box and digit features and input them to an MLP (early fusion), or use late fusion to combine scores of two levels.
Early fusion achieves 43.7 accuracy and shows significant superiority over instance-level method (10.0).
And late fusion achieves a preferable accuracy of 44.2.
Moreover, the part-level method \textit{only} without fusion also obtains an accuracy of 41.4.
This supports our assumption about the effectiveness of part-level representation.

\subsection{Image-based Activity Recognition}
\label{sec:experiment-hico}
Usually, Human-Object Interactions (HOIs) often take up most of the activities, \eg, more than 70\% activities in large-scale datasets~\cite{AVA,activitynet,Kinetics} are HOIs. 
To evaluate \textit{PaSta}Net, we perform image-based HOI recognition on HICO~\cite{hico}.
HICO has 38,116 and 9,658 images in train and test sets and 600 HOIs composed of 117 verbs and 80 COCO objects~\cite{coco}.
Each image has an image-level label which is the aggregation over all HOIs in an image and does not contain any instance boxes.

\noindent{\bf Modes.}
We first pre-train Activity2Vec with \textit{PaSta} labels, then fine-tune Activity2Vec and \textit{PaSta-R} together on HICO train set.
In pre-training and finetuning, 
we \textbf{exclude} the HICO testing data in \textit{PaSta}Net to avoid data pollution.
We 
\begin{table}[H]
	\begin{center}
    \resizebox{0.48\textwidth}{!}{
        \begin{tabular}{lccccc}
		\hline  
	    Method & mAP & Few@1 & Few@5 & Few@10\\
	    \hline  
	    R*CNN~\cite{gkioxari2015contextual} & 28.5 & - & - & -\\
	    Girdhar~\etal~\cite{girdhar2017attentional} & 34.6 & - & - & -\\
	    Mallya~\etal~\cite{Mallya2016Learning} & 36.1 & - & - & -\\
	    Pairwise~\cite{Fang2018Pairwise} & 39.9 & 13.0 & 19.8 & 22.3\\
	    \hline
	    Mallya~\etal~\cite{Mallya2016Learning}+\textit{PaSta}Net*-Linear & 45.0 & 26.5 & 29.1 & 30.3\\ 
	    Pairwise~\cite{Fang2018Pairwise}+\textit{PaSta}Net*-Linear & \textbf{45.9} & 26.2 & 30.6 & 31.8\\
	    Pairwise~\cite{Fang2018Pairwise}+\textit{PaSta}Net*-MLP   & 45.6 & 26.0 & \textbf{30.8} & \textbf{31.9} \\
        Pairwise~\cite{Fang2018Pairwise}+\textit{PaSta}Net*-GCN   & 45.6 & 25.2 & 30.0 & 31.4 \\
        Pairwise~\cite{Fang2018Pairwise}+\textit{PaSta}Net*-Seq   & \textbf{45.9} & 25.3 & 30.2 & 31.6 \\
        Pairwise~\cite{Fang2018Pairwise}+\textit{PaSta}Net*-Tree   & 45.8 & 24.9 & 30.3 & 31.8\\
	    \hline
	    \textit{PaSta}Net*-Linear                               & 44.5 & \textbf{26.9} & 30.0 & 30.7\\
	    Pairwise~\cite{Fang2018Pairwise}+GT-\textit{PaSta}Net*-Linear     & 65.6 & 47.5 & 55.4 & 56.6 \\
        \hline
	    Pairwise~\cite{Fang2018Pairwise}+\textit{PaSta}Net-Linear        & \textbf{46.3} & 24.7 & \textbf{31.8} & \textbf{33.1}\\
	    \hline
		\end{tabular}}
	\end{center}
	\caption{Results on HICO. ``Pairwise~\cite{Fang2018Pairwise}+\textit{PaSta}Net'' means the late fusion of \cite{Fang2018Pairwise} and our part-level result. 
	Few$@$i indicates the mAP on few-shot sets. $@i$ means the number of training images is less than or equal to $i$. 
	The HOI categories number of Few@1, 5, 10 are 49, 125 and 163.
	``\textit{PaSta}Net-x'' means different \textit{PaSta-R}.
	}
	\label{table:all}
\vspace{-0.4cm}
\end{table}
\noindent adopt different data mode to pre-train Activity2Vec:
1) \textbf{``\textit{PaSta}Net*''} mode (38K images): we use the images in HICO train set and their \textit{PaSta} labels. The {\bf only additional supervision} here is the \textit{PaSta} annotations compared to conventional way.
2) \textbf{``GT-\textit{PaSta}Net*''} mode (38K images): the data used is same with \textbf{``\textit{PaSta}Net*''}. To verify the \textbf{upper bound} of our method, we use the ground truth \textit{PaSta} (binary labels) as the predicted \textit{PaSta} probabilities in Activity2Vec. This means we can recognize \textit{PaSta} perfectly and reason out the activities from the best starting point.
3) \textbf{``\textit{PaSta}Net''} mode (118K images): we use all \textit{PaSta}Net images with \textit{PaSta} labels except the HICO testing data.

\noindent{\bf Settings.} We use \textit{image-level PaSta} labels to train Activity2Vec. 
Each image-level \textit{PaSta} label is the aggregation over all existing \textit{PaSta} of all active persons in an image.
For \textit{PaSta} recognition, \ie , we compute the mAP for the \textit{PaSta} categories of each part, and compute the \textit{mean} mAP of all parts.
To be fair, we use the person, body part and object boxes from \cite{Fang2018Pairwise} and VGG-16~\cite{vgg} as the backbone.
The batch size is 16 and the initial learning rate is 1e-5.
We use SGD optimizer with momentum (0.9) and cosine decay restarts~\cite{cosinelr} (the first decay step is 5000).
The pre-training costs 80K iterations and fine-tuning costs 20K iterations.
Image-level \textit{PaSta} and HOI predictions are all generated via Multiple Instance Learning (MIL)~\cite{Maron1998A} of 3 persons and 4 objects.
We choose previous methods~\cite{Mallya2016Learning,Fang2018Pairwise} as the instance-level path in the hierarchical model, and uses late fusion.
Particularly, \cite{Fang2018Pairwise} uses part-pair appearance and location but not part-level \textit{semantics}, thus we still consider it as a baseline to get a more abundant comparison.

\noindent{\bf Results.}
Results are reported in Tab.~\ref{table:all}.
\textbf{\textit{PaSta}Net*} mode methods all outperform the instance-level method. 
The part-level method solely achieves 44.5 mAP and shows good complementarity to the instance-level. 
Their fusion can boost the performance to 45.9 mAP (\textbf{6} mAP improvement). 
And the gap between \cite{Fang2018Pairwise} and \cite{Mallya2016Learning} is largely narrowed from 3.8 to 0.9 mAP.
Activity2Vec achieves \textbf{55.9} mAP on \textit{PaSta} recognition in \textbf{\textit{PaSta}Net*} mode: \textbf{46.3} (head), \textbf{66.8} (arms), \textbf{32.0} (hands), \textbf{68.6} (hip), \textbf{56.2} (thighs), \textbf{65.8} (feet). 
This verifies that \textit{PaSta} can be better learned than activities, thus they can be learned ahead as the basis for reasoning.
In \textbf{GT-\textit{PaSta}Net*} mode, hierarchical paradigm achieves \textbf{65.6} mAP. 
This is a powerful proof of the effectiveness of \textit{PaSta} knowledge. 
Thus what remains to do is to improve the \textit{PaSta} recognition and further promote the activity task performance.
Moreover, in \textbf{\textit{PaSta}Net} mode, we achieve relative \textbf{16}\% improvement.
On few-shot sets, our best result significantly improves \textbf{13.9} mAP, which strongly proves the reusability and transferability of \textit{PaSta}.

\begin{table}
\centering
\resizebox{0.48\textwidth}{!}{
\begin{tabular}{l  c  c  c  c  c  c}
\hline
         & \multicolumn{3}{c}{Default}  &\multicolumn{3}{c}{Known Object} \\
Method         & Full & Rare & Non-Rare  & Full & Rare & Non-Rare \\
\hline
\hline
InteractNet~\cite{Gkioxari2017Detecting} & 9.94  & 7.16  & 10.77     & -   & -   & -\\
GPNN~\cite{qi2018learning}   & 13.11 & 9.34  & 14.23 & - & - & -\\
iCAN~\cite{gao2018ican}      & 14.84 & 10.45 & 16.15 & 16.26 & 11.33 & 17.73\\
TIN~\cite{interactiveness}   & 17.03 & 13.42 & 18.11 & 19.17 & 15.51 & 20.26\\
\hline
iCAN~\cite{gao2018ican}+\textit{PaSta}Net*-Linear    & 19.61 & 17.29 & 20.30 & 22.10 & 20.46 & 22.59\\
TIN~\cite{interactiveness}+\textit{PaSta}Net*-Linear & \textbf{22.12} & \textbf{20.19} & \textbf{22.69} & \textbf{24.06} & \textbf{22.19} & \textbf{24.62} \\
TIN~\cite{interactiveness}+\textit{PaSta}Net*-MLP    & 21.59 & 18.97 & 22.37 & 23.84 & 21.66 & 24.49 \\
TIN~\cite{interactiveness}+\textit{PaSta}Net*-GCN    & 21.73 & 19.55 & 22.38 & 23.95 & 22.14 & 24.49 \\
TIN~\cite{interactiveness}+\textit{PaSta}Net*-Seq    & 21.64 & 19.10 & 22.40 & 23.82 & 21.65 & 24.47 \\
TIN~\cite{interactiveness}+\textit{PaSta}Net*-Tree   & 21.36 & 18.83 & 22.11 & 23.68 & 21.75 & 24.25 \\
\hline
\textit{PaSta}Net*-Linear                            & 19.52 & 17.29 & 20.19 & 21.99 & 20.47 & 22.45 \\
TIN~\cite{interactiveness}+GT-\textit{PaSta}Net*-Linear  & 34.86 & 42.83 & 32.48 & 35.59 & 42.94 & 33.40 \\
\hline
TIN~\cite{interactiveness}+\textit{PaSta}Net-Linear  & \textbf{22.65} & \textbf{21.17} & \textbf{23.09} & \textbf{24.53} & \textbf{23.00} & \textbf{24.99}\\
\hline
\end{tabular}}
\caption{Results on HICO-DET.}
\label{tab:hico-det}
\vspace{-0.6cm}
\end{table}

\subsection{Instance-based Activity Detection}
\label{subsec:exp-HICO-DET}
We further conduct instance-based activity detection on HICO-DET~\cite{hicodet}, which needs to locate human and object and classify the actions simultaneously.
HICO-DET~\cite{hicodet} is a benchmark built on HICO~\cite{hico} and add human and object bounding boxes.  
We choose several state-of-the-arts~\cite{gao2018ican,Gkioxari2017Detecting,qi2018learning,interactiveness} to compare and cooperate. 

\noindent{\bf Settings.}
We use \textit{instance-level PaSta} labels, \ie each annotated person with the corresponding \textit{PaSta} labels, to train Acitivty2Vec, and fine-tune Activity2Vec and \textit{PaSta-R} together on HICO-DET.
All testing data are excluded from pre-training and fine-tining.
We follow the mAP metric of \cite{hicodet}, \ie true positive contains accurate human and object boxes ($IoU>0.5$ with reference to ground truth) and accurate action prediction. 
The metric for \textit{PaSta} detection is similar, \ie, estimated part box and \textit{PaSta} action prediction all have to be accurate.
The mAP of each part and the \textit{mean} mAP are calculated.
For a fair comparison, we use the object detection from \cite{gao2018ican,interactiveness} and ResNet-50\cite{resnet} as backbone. 
We use SGD with momentum (0.9) and cosine decay restart~\cite{cosinelr} (the first decay step is 80K).
The pre-training and fine-tuning take 1M and 2M iterations respectively.
The learning rate is 1e-3 and the ratio of positive and negative samples is 1:4.
A late fusion strategy is adopted.
Three modes in Sec.~\ref{sec:experiment-hico} and different \textit{PaSta-R} are also evaluated.

\noindent{\bf Results.}
Results are shown in Tab.~\ref{tab:hico-det}. 
All \textbf{\textit{PaSta}Net*} mode methods significantly outperform the instance-level methods, which strongly prove the improvement from the learned \textit{PaSta} information.
In \textbf{\textit{PaSta}Net*} mode, the \textit{PaSta} detection performance are 30.2 mAP: 25.8 (head), 44.2 (arms), 17.5 (hands), 41.8 (hip), 22.2 (thighs), 29.9 (feet).
This again verifies that \textit{PaSta} can be well learned.
And \textbf{GT-\textit{PaSta}Net*} (upper bound) and \textbf{\textit{PaSta}Net} (more \textit{PaSta} labels) modes both greatly boosts the performance.
On Rare sets, our method obtains \textbf{7.7} mAP improvement. 

\begin{table}
\centering
\resizebox{0.38\textwidth}{!}{
\begin{tabular}{l c c c}
\hline
Method         &$AP_{role}(Scenario 1)$  & $AP_{role}(Scenario 2)$ \\
\hline
\hline
Gupta \etal~\cite{vcoco}                           & 31.8 & -\\
InteractNet~\cite{Gkioxari2017Detecting}           & 40.0 & -\\
GPNN~\cite{qi2018learning}                         & 44.0 & -\\
iCAN~\cite{gao2018ican}                            & 45.3 & 52.4\\
TIN~\cite{interactiveness}                         & 47.8 & 54.2 \\
\hline
iCAN~\cite{gao2018ican}+\textit{PaSta}Net-Linear       & 49.2 & 55.6\\
TIN~\cite{interactiveness}+\textit{PaSta}Net-Linear    & \textbf{51.0} & \textbf{57.5}\\
\hline
\end{tabular}}
\caption{Transfer learning results on V-COCO~\cite{vcoco}.} 
\label{tab:vcoco}
\vspace{-0.6cm}
\end{table}

\subsection{Transfer Learning with Activity2Vec}
\label{sec:transfer}
To verify the transferability of \textit{PaSta}Net, we design transfer learning experiments on large-scale benchmarks: V-COCO~\cite{vcoco}, HICO-DET~\cite{hicodet} and AVA~\cite{AVA}.
We first use \textit{PaSta}Net to \textbf{pre-train} Activity2Vec and \textit{PaSta-R} with 156 activities and \textit{PaSta} labels.
Then we change the last FC in \textit{PaSta-R} to fit the activity categories of the target benchmark.
Finally, we freeze Activity2Vec and fine-tune \textit{PaSta-R} on the train set of the target dataset.
Here, \textit{PaSta}Net works like the ImageNet~\cite{imagenet} and Activity2Vec is used as a pre-trained knowledge engine to promote other tasks.

\noindent{\bf V-COCO.}
V-COCO contains 10,346 images and instance boxes. 
It has 29 action categories, COCO 80 objects~\cite{coco}.
For a fair comparison, we {\bf exclude} the images of V-COCO and corresponding \textit{PaSta} labels in \textit{PaSta}Net, and use remaining data (109K images) for pre-training. 
We use SGD with 0.9 momenta and cosine decay restarts~\cite{cosinelr} (the first decay is 80K).
The pre-training costs 300K iterations with the learning rate as 1e-3.
The fine-tuning costs 80K iterations with the learning rate as 7e-4.
We select state-of-the-arts~\cite{vcoco,Gkioxari2017Detecting,qi2018learning,gao2018ican,interactiveness} as baselines and adopt the metric $AP_{role}$~\cite{vcoco} (requires accurate human and object boxes and action prediction).
Late fusion strategy is adopted.
With the domain gap, \textit{PaSta}Net still improves the performance by 3.2 mAP (Tab.~\ref{tab:vcoco}.).

\noindent{\bf Image-based AVA.} 
AVA contains 430 video clips with spatio-temporal labels. 
It includes 80 atomic actions consists of body motions and HOIs.
We utilize all \textit{PaSta}Net data (118K images) for pre-training.
Considering that \textit{PaSta}Net is built upon still images, we use the \textbf{ frames per second} as \textit{still images} for \textbf{image-based} instance activity detection. 
We adopt ResNet-50~\cite{resnet} as backbone and SGD with momentum of 0.9. 
The initial learning rate is 1e-2 and the first decay of cosine decay restarts~\cite{cosinelr} is 350K. 
For a fair comparison, we use the human box from \cite{lfb}.
The pre-training costs 1.1M iterations and fine-tuning costs 710K iterations.
We adopt the metric from \cite{AVA}, \ie mAP of the top 60 most common action classes, using IoU threshold of 0.5 between detected human box and the ground truth and accurate action prediction.
For comparison, we adopt a image-based baseline: Faster R-CNN detector~\cite{faster} with
\begin{table}[H]
\begin{center}
\resizebox{0.3\textwidth}{!}{
\begin{tabular}{cc}
\hline  
Method & mAP \\
\hline  
AVA-TF~\cite{ava_tf_baseline}   & 11.4 \\
LFB-Res-50-baseline~\cite{lfb}  & 22.2 \\
LFB-Res-101-baseline~\cite{lfb} & 23.3 \\
\hline
AVA-TF~\cite{ava_tf_baseline}+\textit{PaSta}Net-Linear   & \textbf{15.6} \\
LFB-Res-50-baseline~\cite{lfb}+\textit{PaSta}Net-Linear  & \textbf{23.4} \\ 
LFB-Res-101-baseline~\cite{lfb}+\textit{PaSta}Net-Linear & \textbf{24.3} \\ 
\hline
\end{tabular}}
\end{center}
\caption{Transfer learning results on image-based AVA~\cite{AVA}.}
\label{tab:ava}
\vspace{-0.3cm}
\end{table}
\noindent ResNet-101~\cite{resnet} provided by the AVA website~\cite{ava_tf_baseline}. 
Recent works mainly use a spatial-temporal model such as I3D~\cite{Kinetics}. 
Although \textit{unfair}, we still employ two video-based baselines~\cite{lfb} as instance-level models to cooperate with the part-level method via late fusion. 
Results are listed in Tab.~\ref{tab:ava}. 
Both image and video based methods cooperated with \textit{PaSta}Net achieve impressive improvements, even our model is trained \textbf{without temporal information}. 
Considering the huge domain gap (films) and unseen activities, this result strongly proves its great \textit{generalization ability}.

\noindent{\bf HICO-DET.} 
We exclude the images of HICO-DET and the corresponding \textit{PaSta} labels, and use left data (71K images) for pre-training. 
The test setting in same with Sec.~\ref{subsec:exp-HICO-DET}. 
The pre-training and fine-tuning cost 300K and 1.3M iterations. 
\textit{PaSta}Net shows good transferability and achieve 3.25 mAP improvement on Default Full set (20.28 mAP).

\subsection{Ablation Study}
\label{sec:ablation}
We design ablation studies on HICO-DET with TIN~\cite{interactiveness}+\textit{PaSta*}-Linear (22.12 mAP).
{\bf 1) w/o Part Attention} degrades the performance with 0.21 mAP.
{\bf 2) Language Feature:} We replace the \textit{PaSta} Bert feature in Activity2Vec with: Gaussian noise, Word2Vec~\cite{word2vec} and GloVe~\cite{glove}. 
The results are all worse (20.80, 21.95, 22.01 mAP).
If we change the \textit{PaSta} triplet $\langle part, verb, sth \rangle$ into a sentence and convert it to Bert vector, this vector performs sightly better (22.26 mAP).
This is probably because the sentence carries more contextual information. 
\vspace{-0.3cm}

\section{Conclusion}
In this paper, to make a step toward human activity knowledge engine, we construct \textit{PaSta}Net to provide novel body part-level activity representation (\textit{PaSta}).
Meanwhile, a knowledge transformer Activity2Vec and a part-based reasoning method \textit{PaSta-R} are proposed.
\textit{PaSta}Net brings in interpretability and new possibility for activity understanding. It can effectively bridge the semantic gap between pixels and activities. 
With \textit{PaSta}Net, we significantly boost the performance in supervised and transfer learning tasks, especially under few-shot circumstances.
In the future, we plan to enrich our \textit{PaSta}Net with spatio-temporal \textit{PaSta}. 

{\small
\paragraph{Acknowledgement:} This work is supported in part by the National Key R\&D Program of China, No. 2017YFA0700800, National Natural Science Foundation of China under Grants 61772332 and Shanghai Qi Zhi Institute. 
}

{\small
\bibliographystyle{ieee_fullname}
\bibliography{egbib}

\begin{thebibliography}{10}\itemsep=-1pt

\bibitem{MPII}
Mykhaylo Andriluka, Leonid Pishchulin, Peter Gehler, and Bernt Schiele.
\newblock 2d human pose estimation: New benchmark and state of the art analysis.
\newblock In {\em CVPR}, 2014.

\bibitem{Kinetics}
Joao Carreira and Andrew Zisserman.
\newblock Quo vadis, action recognition? a new model and the kinetics dataset.
\newblock In {\em CVPR}, 2017.

\bibitem{hico}
Yu~Wei Chao, Zhan Wang, Yugeng He, Jiaxuan Wang, and Jia Deng.
\newblock Hico: A benchmark for recognizing human-object interactions in images.
\newblock In {\em ICCV}, 2015.

\bibitem{hicodet}
Yu-Wei Chao, Yunfan Liu, Xieyang Liu, Huayi Zeng, and Jia Deng.
\newblock Learning to detect human-object interactions.
\newblock In {\em WACV}, 2018.

\bibitem{choutas2018potion}
Vasileios Choutas, Philippe Weinzaepfel, J{\'e}r{\^o}me Revaud, and Cordelia Schmid.
\newblock Potion: Pose motion representation for action recognition.
\newblock In {\em CVPR}, 2018.

\bibitem{church1990word}
Kenneth~Ward Church and Patrick Hanks.
\newblock Word association norms, mutual information, and lexicography.
\newblock In {\em Computational linguistics}, 1990.

\bibitem{Delaitre2011Learning}
Vincent Delaitre, Josef Sivic, and Ivan Laptev.
\newblock Learning person-object interactions for action recognition in still images.
\newblock In {\em NIPS}, 2011.

\bibitem{imagenet}
Jia Deng, Wei Dong, Richard Socher, Li-Jia Li, Kai Li, and Li~Fei-Fei.
\newblock Imagenet: A large-scale hierarchical image database.
\newblock In {\em CVPR}, 2009.

\bibitem{devlin2018bert}
Jacob Devlin, Ming-Wei Chang, Kenton Lee, and Kristina Toutanova.
\newblock Bert: Pre-training of deep bidirectional transformers for language understanding.
\newblock In {\em arXiv preprint arXiv:1810.04805}, 2018.

\bibitem{Reductionism}
Wendy Doniger.
\newblock In {\em Reductionism}.
\newblock 1999.

\bibitem{Du_2015_CVPR}
Yong Du, Wei Wang, and Liang Wang.
\newblock Hierarchical recurrent neural network for skeleton based action recognition.
\newblock In {\em CVPR}, 2015.

\bibitem{activitynet}
Bernard~Ghanem Fabian Caba~Heilbron, Victor~Escorcia and Juan~Carlos Niebles.
\newblock Activitynet: A large-scale video benchmark for human activity understanding.
\newblock In {\em CVPR}, 2015.

\bibitem{fang2017rmpe}
Hao-Shu Fang, Shuqin Xie, Yu-Wing Tai, and Cewu Lu.
\newblock {RMPE}: Regional multi-person pose estimation.
\newblock In {\em ICCV}, 2017.

\bibitem{Fang2018Pairwise}
Hao~Shu Fang, Jinkun Cao, Yu~Wing Tai, and Cewu Lu.
\newblock Pairwise body-part attention for recognizing human-object interactions.
\newblock In {\em ECCV}, 2018.

\bibitem{fang2019instaboost}
Hao-Shu Fang, Jianhua Sun, Runzhong Wang, Minghao Gou, Yong-Lu Li, and Cewu Lu.
\newblock Instaboost: Boosting instance segmentation via probability map guided copy-pasting.
\newblock In {\em ICCV}, 2019.

\bibitem{Feichtenhofer_2016_CVPR}
Christoph Feichtenhofer, Axel Pinz, and Andrew Zisserman.
\newblock Convolutional two-stream network fusion for video action recognition.
\newblock In {\em CVPR}, 2016.

\bibitem{gao2018ican}
Chen Gao, Yuliang Zou, and Jia-Bin Huang.
\newblock ican: Instance-centric attention network for human-object interaction detection.
\newblock In {\em arXiv preprint arXiv:1808.10437}, 2018.

\bibitem{girdhar2017attentional}
Rohit Girdhar and Deva Ramanan.
\newblock Attentional pooling for action recognition.
\newblock In {\em NIPS}, 2017.

\bibitem{Gkioxari2014R}
Georgia Gkioxari, Bharath Hariharan, Ross Girshick, and Jitendra Malik.
\newblock R-cnns for pose estimation and action detection.
\newblock In {\em Computer Science}, 2014.

\bibitem{Gkioxari_2015_ICCV}
Georgia Gkioxari, Ross Girshick, and Jitendra Malik.
\newblock Actions and attributes from wholes and parts.
\newblock In {\em ICCV}, 2015.

\bibitem{gkioxari2015contextual}
Georgia Gkioxari, Ross Girshick, and Jitendra Malik.
\newblock Contextual action recognition with r* cnn.
\newblock In {\em ICCV}, 2015.

\bibitem{Gkioxari2017Detecting}
Georgia Gkioxari, Ross Girshick, Piotr Doll{\'a}r, and Kaiming He.
\newblock Detecting and recognizing human-object interactions.
\newblock In {\em CVPR}, 2018.

\bibitem{AVA}
Chunhui Gu, Chen Sun, David~A Ross, Carl Vondrick, Caroline Pantofaru, Yeqing Li, Sudheendra Vijayanarasimhan, George Toderici, Susanna Ricco, Rahul Sukthankar, et~al.
\newblock Ava: A video dataset of spatio-temporally localized atomic visual actions.
\newblock In {\em CVPR}, 2018.

\bibitem{ava_tf_baseline}
Chunhui Gu, Chen Sun, David~A Ross, Carl Vondrick, Caroline Pantofaru, Yeqing Li, Sudheendra Vijayanarasimhan, George Toderici, Susanna Ricco, Rahul Sukthankar, et~al.
\newblock Ava.
\newblock \url{https://research.google.com/ava/download.html}, 2018.

\bibitem{vcoco}
Saurabh Gupta and Jitendra Malik.
\newblock Visual semantic role labeling.
\newblock In {\em arXiv preprint arXiv:1505.04474}, 2015.

\bibitem{resnet}
Kaiming He, Xiangyu Zhang, Shaoqing Ren, and Jian Sun.
\newblock Deep residual learning for image recognition.
\newblock In {\em CVPR}, 2016.

\bibitem{he2017mask}
Kaiming He, Georgia Gkioxari, Piotr Doll{\'a}r, and Ross Girshick.
\newblock Mask r-cnn.
\newblock In {\em ICCV}, 2017.

\bibitem{lstm}
Sepp Hochreiter and J{\"u}rgen Schmidhuber.
\newblock Long short-term memory.
\newblock In {\em Neural computation}, 1997.

\bibitem{Hu2014Recognising}
Jian~Fang Hu, Wei~Shi Zheng, Jianhuang Lai, Shaogang Gong, and Tao Xiang.
\newblock Recognising human-object interaction via exemplar based modelling.
\newblock In {\em ICCV}, 2014.

\bibitem{6165309}
Shuiwang Ji, Wei Xu, Ming Yang, and Kai Yu.
\newblock 3d convolutional neural networks for human action recognition.
\newblock In {\em TPAMI}, 2012.

\bibitem{gcn}
Thomas~N Kipf and Max Welling.
\newblock Semi-supervised classification with graph convolutional networks.
\newblock In {\em arXiv preprint arXiv:1609.02907}, 2016.

\bibitem{openimages}
Ivan Krasin, Tom Duerig, Neil Alldrin, Vittorio Ferrari, Sami Abu-El-Haija, Alina Kuznetsova, Hassan Rom, Jasper Uijlings, Stefan Popov, Andreas Veit, et~al.
\newblock Openimages: A public dataset for large-scale multi-label and multi-class image classification.
\newblock In {\em Dataset available from https://github. com/openimages}, 2017.

\bibitem{interactiveness}
Yong-Lu Li, Siyuan Zhou, Xijie Huang, Liang Xu, Ze~Ma, Hao-Shu Fang, Yanfeng Wang, and Cewu Lu.
\newblock Transferable interactiveness knowledge for human-object interaction detection.
\newblock In {\em CVPR}, 2019.

\bibitem{coco}
Tsung~Yi Lin, Michael Maire, Serge Belongie, James Hays, Pietro Perona, Deva Ramanan, Piotr Dollár, and C.~Lawrence Zitnick.
\newblock Microsoft coco: Common objects in context.
\newblock In {\em ECCV}, 2014.

\bibitem{pic}
Si~Liu, Lejian Ren, Yue Liao, Guanghui Ren, Hongyi Xiang, and Guanbin Li.
\newblock Pic-person in context.
\newblock \url{http://picdataset.com/challenge/index/}, 2018.

\bibitem{cosinelr}
Ilya Loshchilov and Frank Hutter.
\newblock Sgdr: Stochastic gradient descent with warm restarts.
\newblock In {\em arXiv preprint arXiv:1608.03983}, 2016.

\bibitem{Lu2016Visual}
Cewu Lu, Ranjay Krishna, Michael Bernstein, and Fei~Fei Li.
\newblock Visual relationship detection with language priors.
\newblock In {\em ECCV}, 2016.

\bibitem{partstate}
Cewu Lu, Hao Su, Yonglu Li, Yongyi Lu, Li~Yi, Chi-Keung Tang, and Leonidas~J Guibas.
\newblock Beyond holistic object recognition: Enriching image understanding with part states.
\newblock In {\em CVPR}, 2018.

\bibitem{tsne}
Laurens van~der Maaten and Geoffrey Hinton.
\newblock Visualizing data using t-sne.
\newblock In {\em JMLR}, 2008.

\bibitem{Maji2012Action}
S.~Maji, L.~Bourdev, and J.~Malik.
\newblock Action recognition from a distributed representation of pose and appearance.
\newblock In {\em CVPR}, 2011.

\bibitem{Mallya2016Learning}
Arun Mallya and Svetlana Lazebnik.
\newblock Learning models for actions and person-object interactions with transfer to question answering.
\newblock In {\em ECCV}, 2016.

\bibitem{Maron1998A}
Oded Maron and Tomás Lozano-Pérez.
\newblock A framework for multiple-instance learning.
\newblock In {\em NIPS}, 1998.

\bibitem{word2vec}
Tomas Mikolov, Kai Chen, Greg Corrado, and Jeffrey Dean.
\newblock Efficient estimation of word representations in vector space.
\newblock In {\em arXiv preprint arXiv:1301.3781}, 2013.

\bibitem{miller1995wordnet}
George~A Miller.
\newblock Wordnet: a lexical database for english.
\newblock In {\em Communications of the ACM}, 1995.

\bibitem{pang2019deep}
Bo~Pang, Kaiwen Zha, Hanwen Cao, Chen Shi, and Cewu Lu.
\newblock Deep rnn framework for visual sequential applications.
\newblock In {\em CVPR}, 2019.

\bibitem{pang2020deep}
Bo~Pang, Kaiwen Zha, Yifan Zhang, and Cewu Lu.
\newblock Further understanding videos through adverbs: A new video task.
\newblock In {\em AAAI}, 2020.

\bibitem{glove}
Jeffrey Pennington, Richard Socher, and Christopher Manning.
\newblock Glove: Global vectors for word representation.
\newblock In {\em EMNLP}, 2014.

\bibitem{qi2018learning}
Siyuan Qi, Wenguan Wang, Baoxiong Jia, Jianbing Shen, and Song-Chun Zhu.
\newblock Learning human-object interactions by graph parsing neural networks.
\newblock In {\em ECCV}, 2018.

\bibitem{faster}
Shaoqing Ren, Kaiming He, Ross Girshick, and Jian Sun.
\newblock Faster r-cnn: Towards real-time object detection with region proposal networks.
\newblock In {\em NIPS}, 2015.

\bibitem{shao2018find}
Dian Shao, Yu~Xiong, Yue Zhao, Qingqiu Huang, Yu~Qiao, and Dahua Lin.
\newblock Find and focus: Retrieve and localize video events with natural language queries.
\newblock In {\em ECCV}, 2018.

\bibitem{NIPS2014_5353}
Karen Simonyan and Andrew Zisserman.
\newblock Two-stream convolutional networks for action recognition in videos.
\newblock In {\em NIPS}, 2014.

\bibitem{vgg}
Karen Simonyan and Andrew Zisserman.
\newblock Very deep convolutional networks for large-scale image recognition.
\newblock In {\em arXiv preprint arXiv:1409.1556}, 2014.

\bibitem{UCF101}
Khurram Soomro, Amir~Roshan Zamir, and Mubarak Shah.
\newblock Ucf101: A dataset of 101 human actions classes from videos in the wild.
\newblock In {\em arXiv preprint arXiv:1212.0402}, 2012.

\bibitem{Sun_2015_ICCV}
Lin Sun, Kui Jia, Dit-Yan Yeung, and Bertram~E. Shi.
\newblock Human action recognition using factorized spatio-temporal convolutional networks.
\newblock In {\em ICCV}, 2015.

\bibitem{Thurau2008Pose}
Christian Thurau and Vaclav Hlavac.
\newblock Pose primitive based human action recognition in videos or still images.
\newblock In {\em CVPR}, 2008.

\bibitem{Tran_2015_ICCV}
Du~Tran, Lubomir Bourdev, Rob Fergus, Lorenzo Torresani, and Manohar Paluri.
\newblock Learning spatiotemporal features with 3d convolutional networks.
\newblock In {\em ICCV}, 2015.

\bibitem{Vemulapalli_2014_CVPR}
Raviteja Vemulapalli, Felipe Arrate, and Rama Chellappa.
\newblock Human action recognition by representing 3d skeletons as points in a lie group.
\newblock In {\em CVPR}, 2014.

\bibitem{vinyals2015show}
Oriol Vinyals, Alexander Toshev, Samy Bengio, and Dumitru Erhan.
\newblock Show and tell: A neural image caption generator.
\newblock In {\em CVPR}, 2015.

\bibitem{lfb}
Chao-Yuan Wu, Christoph Feichtenhofer, Haoqi Fan, Kaiming He, Philipp Krahenbuhl, and Ross Girshick.
\newblock Long-term feature banks for detailed video understanding.
\newblock In {\em CVPR}, 2019.

\bibitem{xu2018srda}
Wenqiang Xu, Yonglu Li, and Cewu Lu.
\newblock Srda: Generating instance segmentation annotation via scanning, reasoning and domain adaptation.
\newblock In {\em ECCV}, 2018.

\bibitem{Yang2010Recognizing}
Weilong Yang, Yang Wang, and Greg Mori.
\newblock Recognizing human actions from still images with latent poses.
\newblock In {\em CVPR}, 2010.

\bibitem{Yao2010Modeling}
Bangpeng Yao and Fei~Fei Li.
\newblock Modeling mutual context of object and human pose in human-object interaction activities.
\newblock In {\em CVPR}, 2010.

\bibitem{Yao2012Action}
Bangpeng Yao and Fei~Fei Li.
\newblock Action recognition with exemplar based 2.5d graph matching.
\newblock In {\em ECCV}, 2012.

\bibitem{Yao2011Human}
Bangpeng Yao, Xiaoye Jiang, Aditya Khosla, Andy~Lai Lin, Leonidas Guibas, and Fei~Fei Li.
\newblock Human action recognition by learning bases of action attributes and parts.
\newblock In {\em ICCV}, 2011.

\bibitem{zhao2017single}
Zhichen Zhao, Huimin Ma, and Shaodi You.
\newblock Single image action recognition using semantic body part actions.
\newblock In {\em ICCV}, 2017.

\bibitem{hcvrd}
Bohan Zhuang, Qi~Wu, Chunhua Shen, Ian Reid, and Anton van~den Hengel.
\newblock Care about you: towards large-scale human-centric visual relationship detection.
\newblock In {\em arXiv preprint arXiv:1705.09892}, 2017.

\bibitem{JhuangICCV2013}
H. Jhuang and J. Gall and S. Zuffi and C. Schmid and M. J. Black.
\newblock Towards understanding action recognition.
\newblock In {\em ICCV}, 2013.

\end{thebibliography}
}

\clearpage

\onecolumn
\begin{appendices}

\begin{figure*}[!ht]
	\begin{center}
        \begin{minipage}{.45\linewidth}
            \centerline{\includegraphics[width=\linewidth]{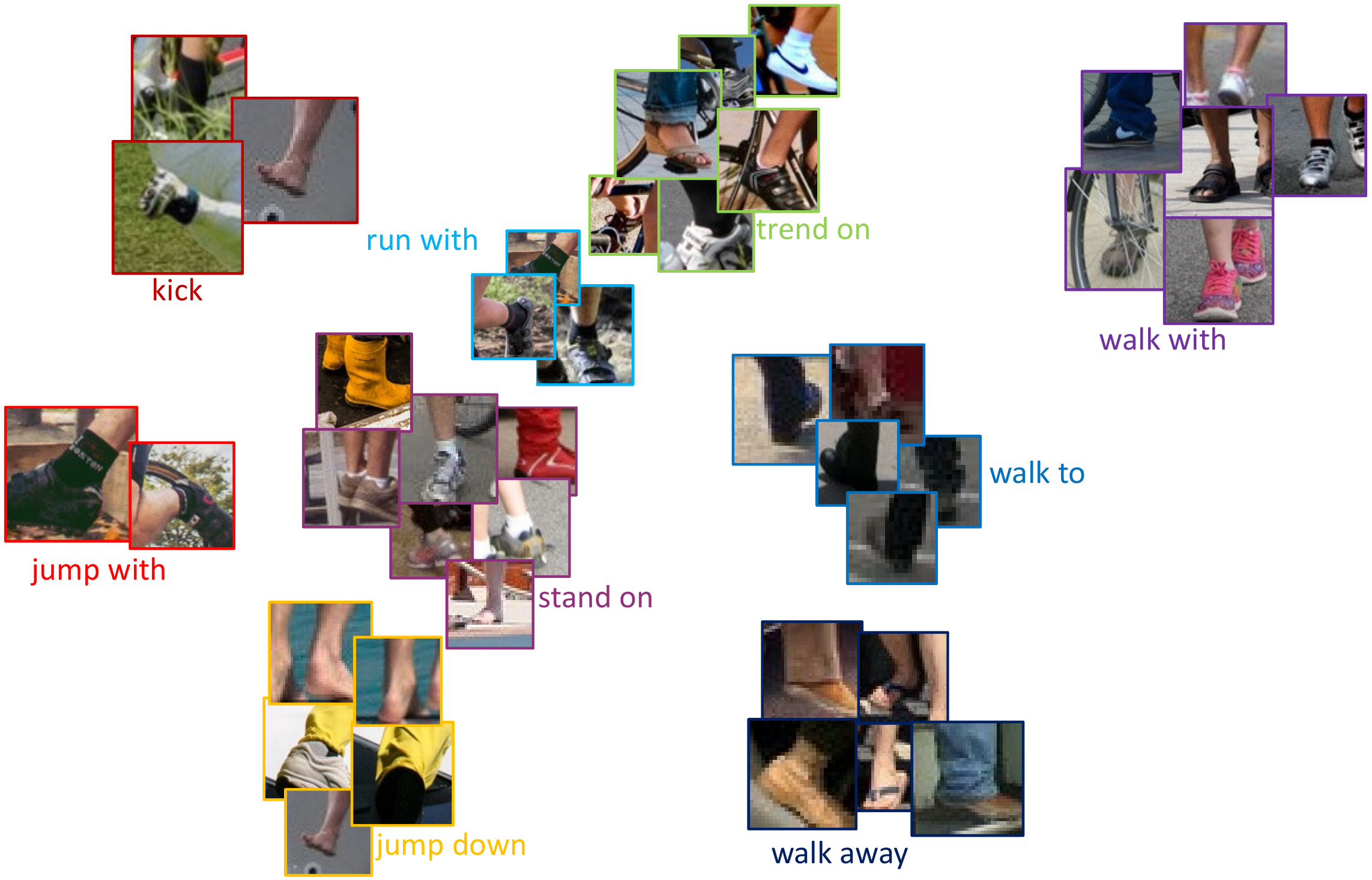}}
            \centerline{(a) \textit{PaSta} of foot}
        \end{minipage}
        \hfill
        \begin{minipage}{.48\linewidth}
            \centerline{\includegraphics[width=\linewidth]{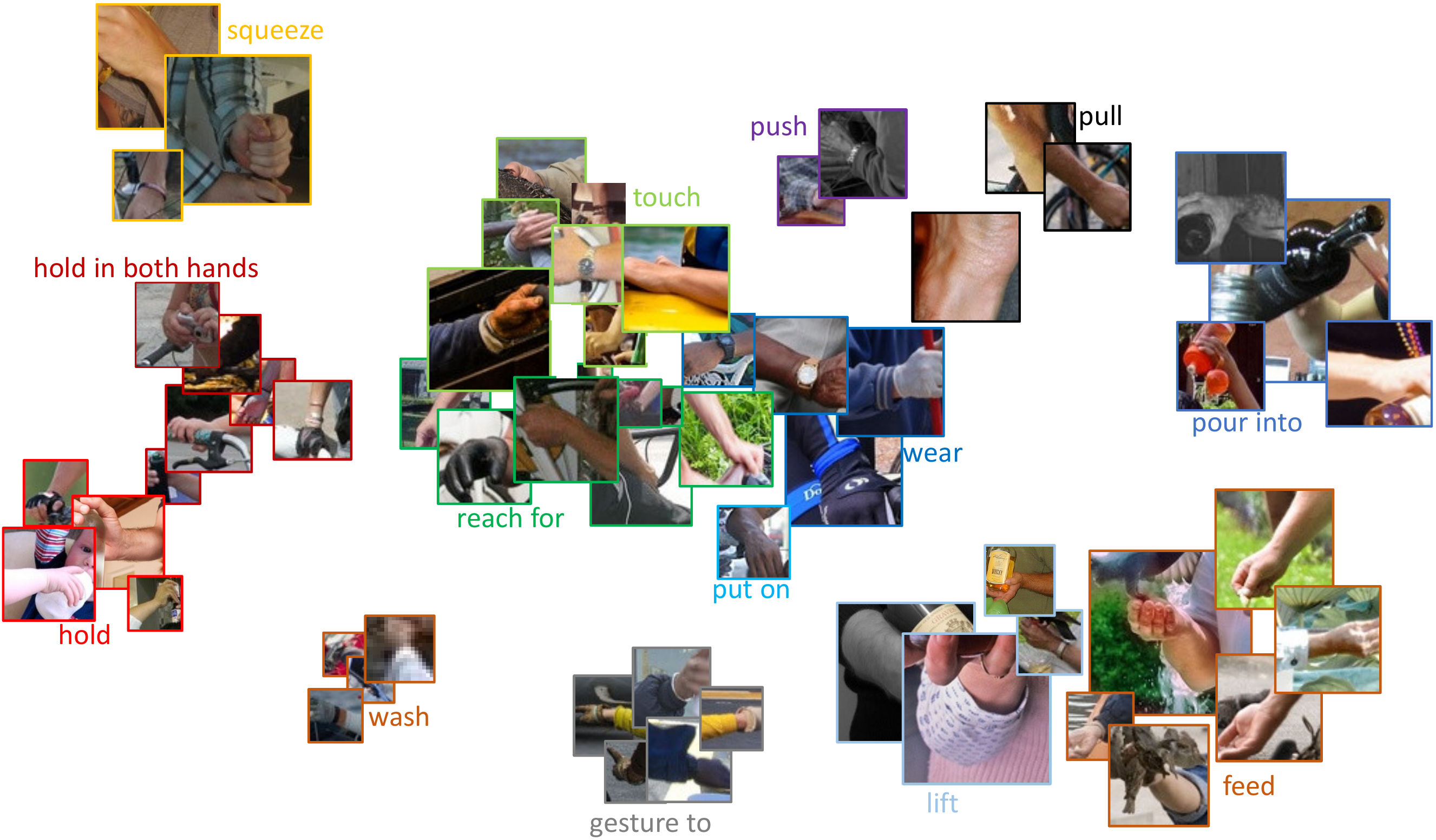}}
            \centerline{(b) \textit{PaSta} of hand}
        \end{minipage}
	\end{center}
	\caption{Visualized \textit{PaSta} representations via t-SNE~\cite{tsne}. Sub-figure (a) depicts the \textit{PaSta} features of foot and sub-figure (b) depicts \textit{PaSta} features of hand. Our model can extract meaningful and reasonable features for various \textit{PaSta}.}
	\label{Figure:t-SNE} 
\end{figure*}

\begin{table*}[!ht]
 	\begin{center}
 		\begin{tabular}{ccccccc}
 		\hline  
 	    Image & Human-Instance& Object-Instance & Activity & \textit{PaSta} & Existing-Datasets & Crowdsourcing\\
 	    \hline  
 	    118,995 & 285,921 & 250,621 & 724,855 & 7,248,550 & 95,027 & 23,968\\
 	    \hline
 		\end{tabular}
 	\end{center}
 	\caption{Characteristics of \textit{Pasta}Net. ``Existing-Datasets'' and ``Crowdsourcing'' indicate the image sources of \textit{Pasta}Net.}
 	\label{table:statistics}
\end{table*}

\begin{figure*}[!ht]
 	\begin{center}
 		\includegraphics[width=1\textwidth]{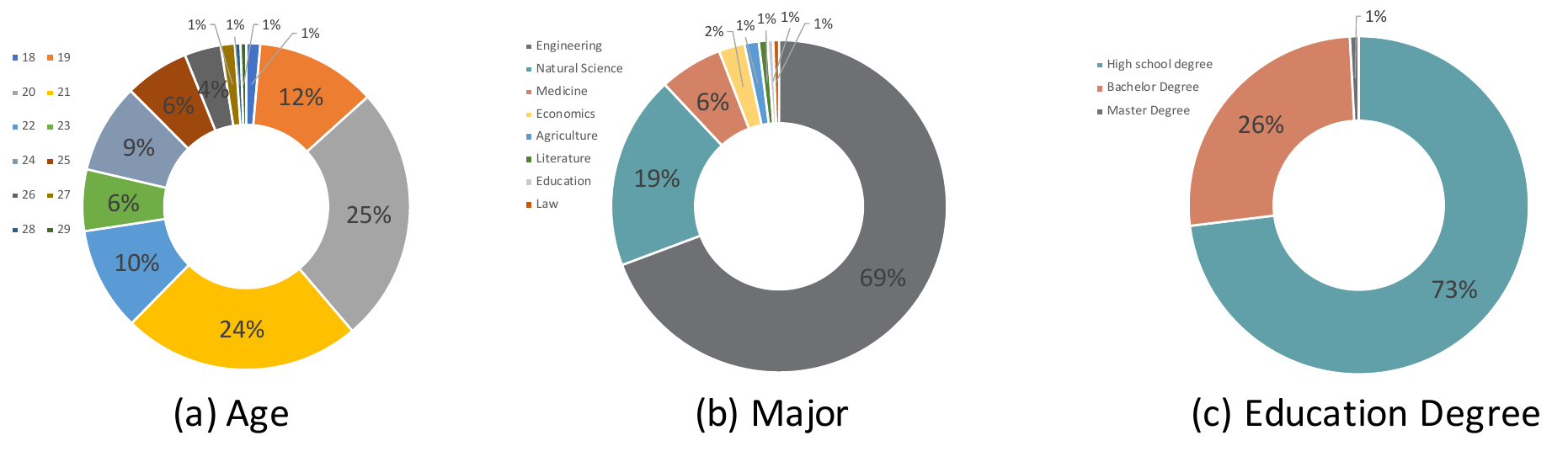}
 	\end{center}
 	\caption{Annotator backgrounds.}
 	\label{Figure:background}
\end{figure*}

\section{Dataset Details}
In this section, we give a more detailed introduction of our knowledge base \textit{PaSta}Net, covering characteristics and the annotator backgrounds.

\subsection{Characteristics of \textit{Pasta}Net}
Tab.~\ref{table:statistics} shows some characteristics of \textit{PaSta}Net. Now \textit{PaSta}Net contains about 118K+ images and the corresponding instance-level and human part-level annotations.
To give an intuitive presentation of our \textit{PaSta}, we use t-SNE~\cite{tsne} to visualize the part-level features of some typical \textit{PaSta} samples in Fig.~\ref{Figure:t-SNE}.
We use the human body part patches with different colored borders to replace the embedding points in Fig.~\ref{Figure:t-SNE}, \eg the embeddings of ``hand holds something'' and ``foot kicks something''.

\begin{table*}[htb]
  \centering
  \begin{tabular}{p{60pt}p{410pt}}
    \toprule    
    HOIs    & adjust, assemble, block, blow, board, break, brush with, board gaming, buy, carry, catch, chase, check, chop, clean, clink glass, close, control, cook, cut, cut with, dig, direct, drag, dribble, drink with, drive, dry, eat, eat at, enter, exit, extract, feed, fill, flip, flush, fly, fight, fishing, give sth to sb, grab, greet, grind, groom, hand shake, herd, hit, hold, hop on, hose, hug, hunt, inspect, install, jump, kick, kiss, lasso, launch, lick, lie on, lift, light, listen to sth, listen to a person, load, lose, make, milk, move, open, operate, pack, paint, park, pay, peel, pet, play musical instrument, play with sb, play with pets, pick, pick up, point, pour, press, pull, push, put down, put on, race, read, release, repair, ride, row, run, sail, scratch, serve, set, shear, shoot, shovel, sign, sing to sb, sip, sit at, sit on, slide, smell, smoke, spin, squeeze, stab, stand on, stand under, stick, stir, stop at, straddle, swing, tag, take a photo, take sth from sb, talk on, talk to, teach, text on, throw, tie, toast, touch, train, turn, type on, walk, wash, watch, wave, wear, wield, work on laptop, write, zip\\ 
    \midrule
    Body Motions           & bow, clap, climb, crawl, dance, fall, get up, kneel, physical exercise, swim\\
    \midrule
    \midrule
    Objects & airplane, apple, backpack, banana, baseball bat, baseball glove, bear, bed, bench, bicycle, bird, boat, book, bottle, bowl, broccoli, bus, cake, car, carrot, cat, cell phone, chair, clock, couch, cow, cup, dining table, dog, donut, elephant, fire hydrant, fork, frisbee, giraffe, hair drier, handbag, horse, hot dog, keyboard, kite, knife, laptop, microwave, motorcycle, mouse, orange, oven, parking meter, person, pizza, potted plant, refrigerator, remote, sandwich, scissors, sheep, sink, skateboard, skis, snowboard, spoon, sports ball, stop sign, suitcase, surfboard, teddy bear, tennis racket, tie, toaster, toilet, toothbrush, traffic light, train, truck, tv, umbrella, vase, wine glass, zebra\\
    \bottomrule
  \end{tabular}
  \caption{Activities and related objects in our \textit{PaSta}Net. HOIs indicate the interactions between person and object/person.}
  \label{tab:action}
\end{table*}
\begin{table*}[htb]
  \centering
  \begin{tabular}{p{60pt}p{410pt}}
    \toprule    
    Head    & eat, inspect, talk with sth, talk to, close with, kiss, raise up, lick, blow, drink with, smell, wear, no action\\ 
    \midrule
    Arm  & carry, close to, hug, swing, no action  \\ 
    \midrule
    Hand & hold, carry, reach for, touch, put on, twist, wear, throw, throw out, write on, point with, point to, use sth to point to, press, squeeze, scratch, pinch, gesture to, push, pull, pull with sth, wash, wash with sth, hold in both hands, lift, raise, feed, cut with sth, catch with sth, pour into, no action \\
    \midrule
    Hip & sit on, sit in, sit beside, close with, no action \\
    \midrule
    Thigh & walk with, walk to, run with, run to, jump with, close with, straddle, jump down, walk away, no action\\ 
    \midrule
    Foot & stand on, step on, walk with, walk to, run with, run to, dribble, kick, jump down, jump with, walk away, no action \\ 
    \bottomrule
  \end{tabular}
  \caption{Human body part states (\textit{PaSta}) in our \textit{PaSta}Net.}
  \label{tab:pasta}
\end{table*}

\subsection{Annotator Backgrounds}
There are about 360 annotators have participated in the construction of \textit{Pasta}Net. 
They have various backgrounds, thus we can ensure annotation diversity and reduce the bias. The specific information is shown in Fig.~\ref{Figure:background}.
 
\section{Selected Activities and \textit{PaSta}}
\begin{figure*}[!ht]
 	\begin{center}
 		\includegraphics[width=0.7\textwidth]{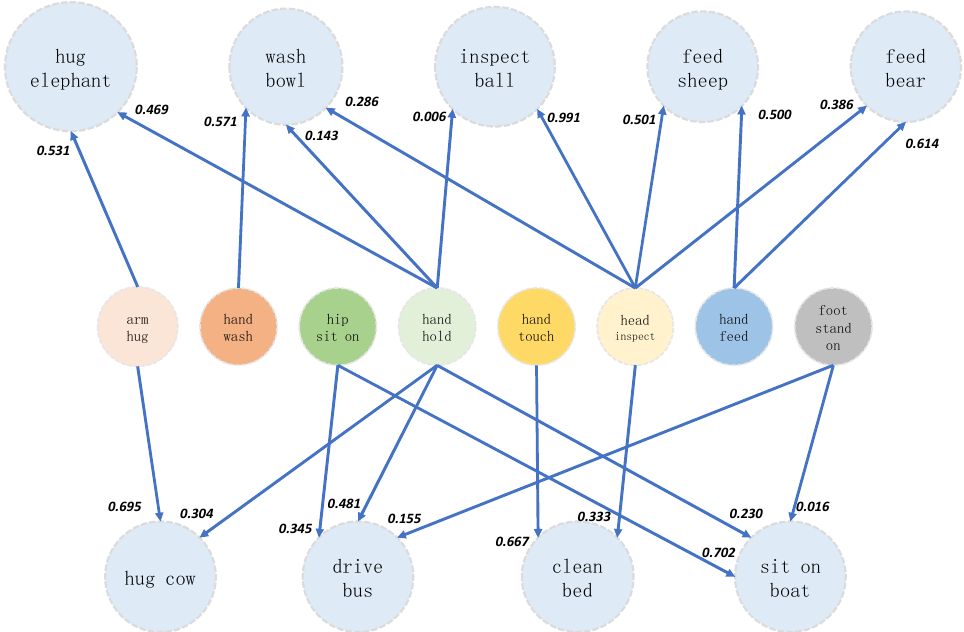}
 	\end{center}
 	\caption{Part of the Activity Parsing Tree.}
 	\label{Figure:rel_part}
 \end{figure*}
 
\begin{figure*}[!ht]
 	\begin{center}
 		\includegraphics[width=0.95\textwidth]{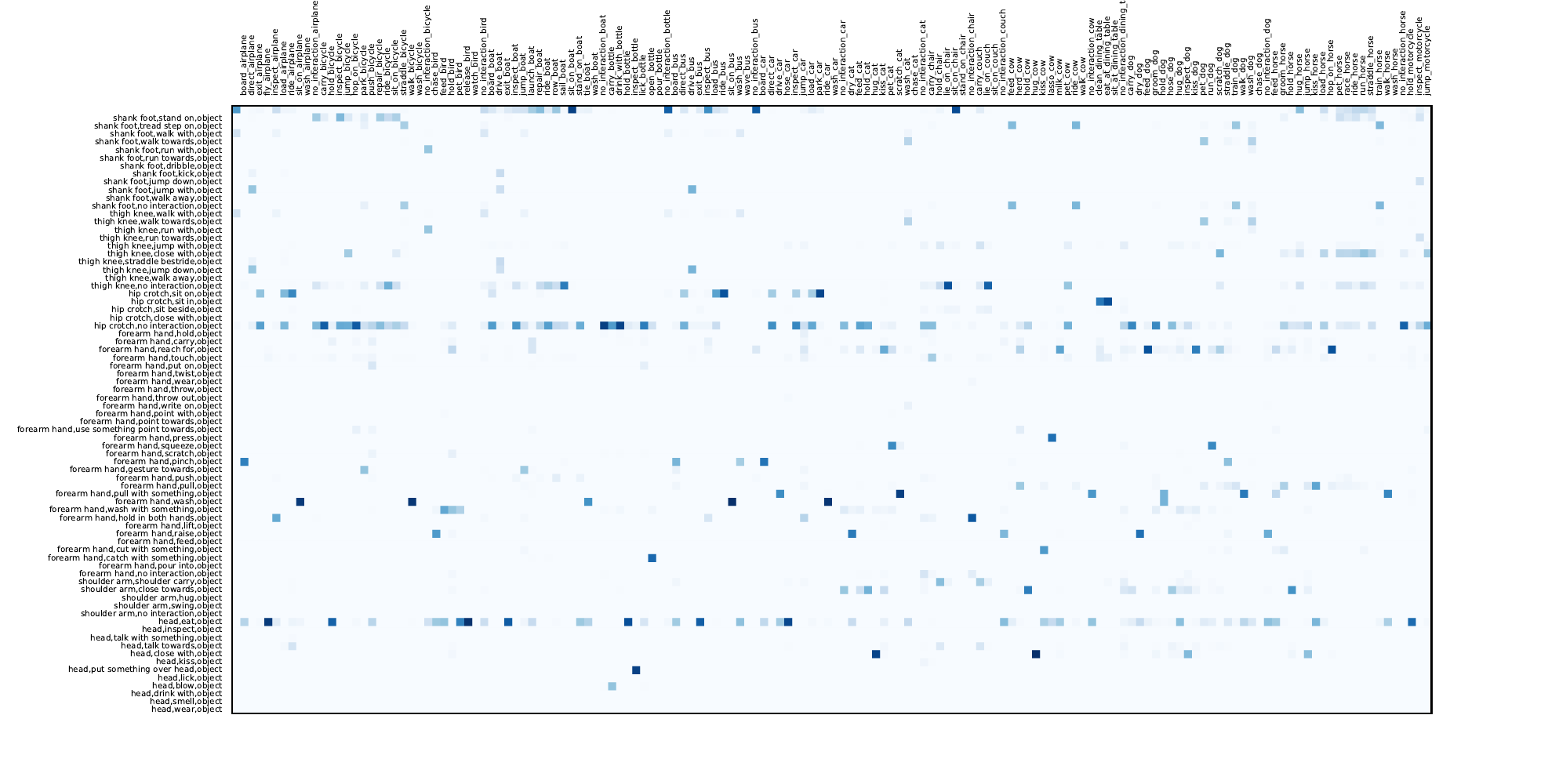}
 	\end{center}
 	\caption{Part of the Activity-\textit{Pasta} Co-occurrence matrix.}
 	\label{Figure:rel_part1}
 \end{figure*}
We list all selected 156 activities and 76 \textit{PaSta} in Tab.~\ref{tab:action} and Tab.~\ref{tab:pasta}.
\textit{PaSta}Net contains both person-object/person interactions and body only motions, which cover the vast majority of human daily life activities.
And all the annotated interacted objects in interactions all belong to COCO 80 object categories~\cite{coco}.

\subsection{Activity and \textit{PaSta}}
\textit{PaSta}Net can provide abundant activity knowledge for both instance and part levels and help construct a large-scale activity parsing tree, as seen in Fig.~\ref{Figure:rel_part}.
Moreover, we can represent the parsing tree as a co-occurrence matrix of the activities and \textit{PaSta}. A part of the matrix is depicted in Fig.~\ref{Figure:rel_part1}.

\begin{figure*}[!ht]
 	\begin{center}
 		\includegraphics[width=0.8\textwidth]{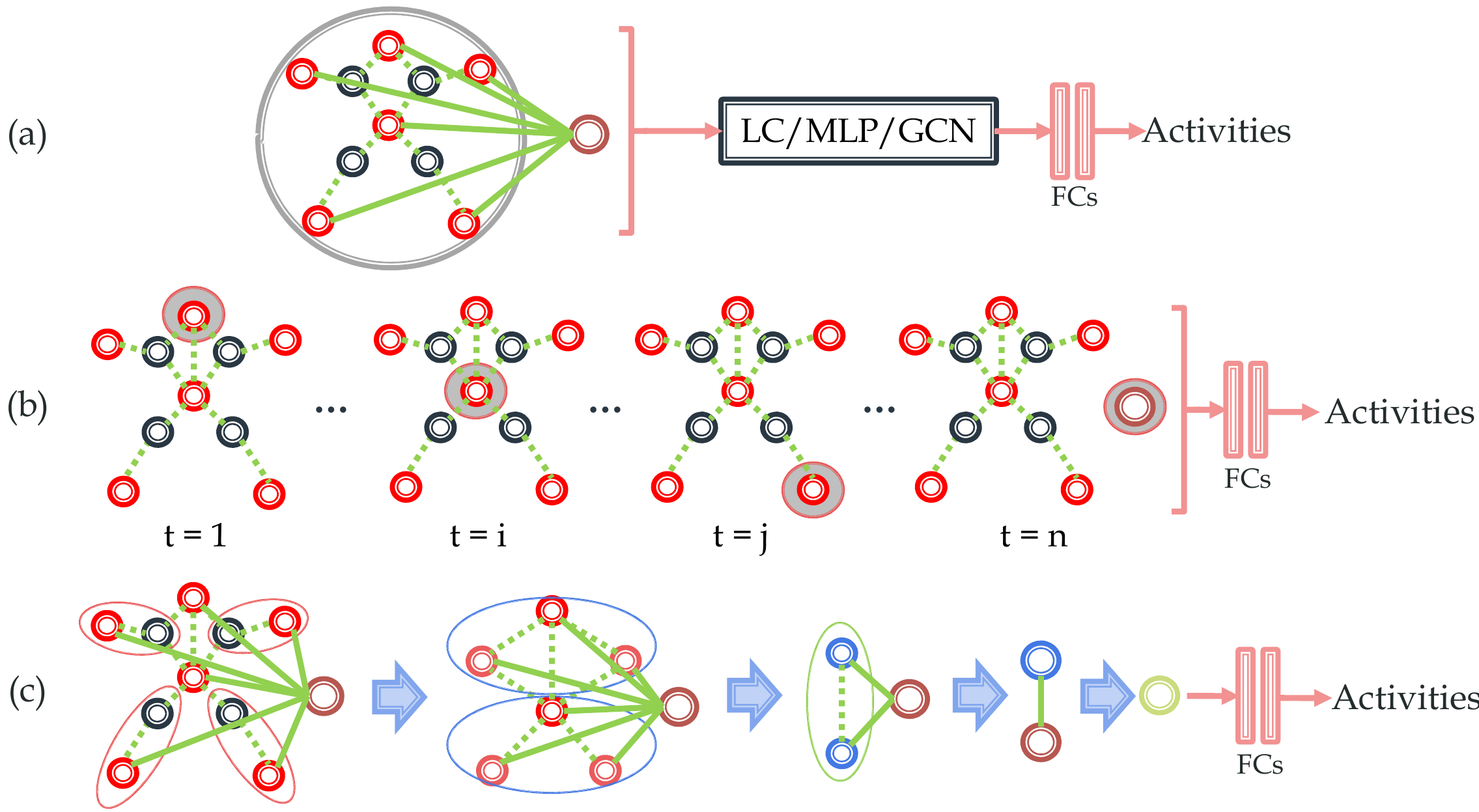}
 	\end{center}
 	\caption{\textit{PaSta-R} model: (a) Linear Combination, MLP and GCN; (b) Sequential Model; (c) Tree Structured Passing.}
 	\label{Figure:reasoning-model}
\end{figure*}

\begin{figure*}[!ht]
 	\begin{center}
 		\includegraphics[width=0.8\textwidth]{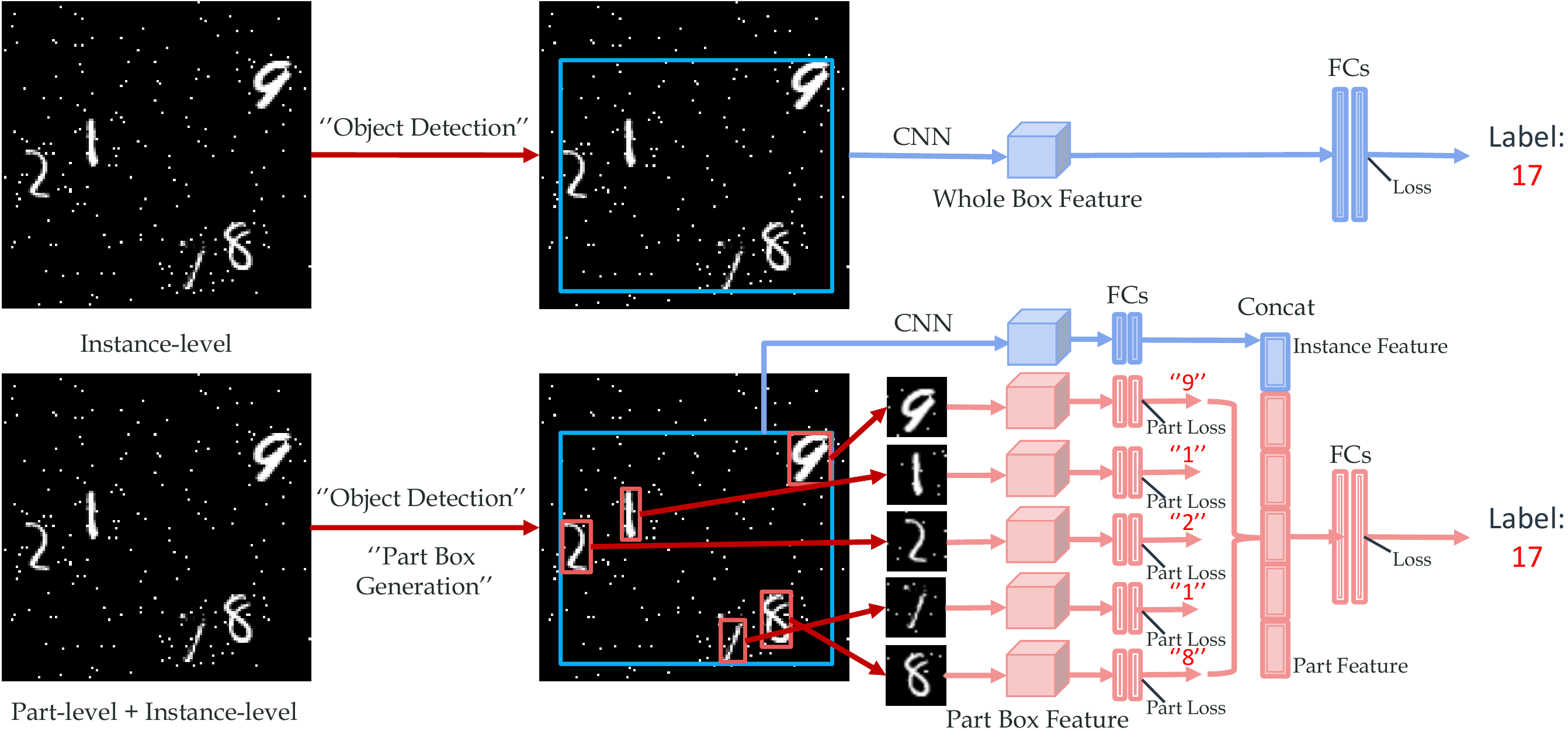}
 	\end{center}
 	\caption{Instance-based and hierarchical models in MNIST-Action.}
 	\label{Figure:mnist}
\end{figure*}

\section{Additional Details of \textit{PaSta}-R}
Fig.~\ref{Figure:reasoning-model} gives more details about the different \textit{PaSta}-R implementations, \ie directly input the Activity2Vec output to the Linear Combination, MLP or GCN~\cite{gcn}, sequential LSTM-based model and the tree-structured passing model.
 
\section{Additional Details of MNIST-Action}
In this section, we provide some details of the  MNIST-Action experiment.
The instance-based and hierarchical models are shown in Fig.~\ref{Figure:mnist}.
The train and test set sizes are 5,000 and 800.
In the instance-based model, we directly use the ROI pooling feature of the digit union box to predict the target (summation of the largest and second largest numbers).
We use optimizer RMSProp and the initial learning rate is 0.0001. The batch size is 16 and we train the model 1K epochs.
In the hierarchical model, we first operate the single digit recognition and then use single digit features (part features) together with the instance feature to infer the sum (early fusion). If using late fusion, we directly fuse the scores of instance branch and part branch.
We also use optimizer RMSProp and the initial learning rate is 0.001. The batch size is 32 and the training also costs 1K epochs.
Two models are all implemented with 4 convolution layers with subsequent fully-connect layers.
\begin{figure*} [!ht]
   \centering 
   \subfigure[Loss]{ 
     \label{Figure:exp1_loss} 
     \includegraphics[width=0.45\textwidth]{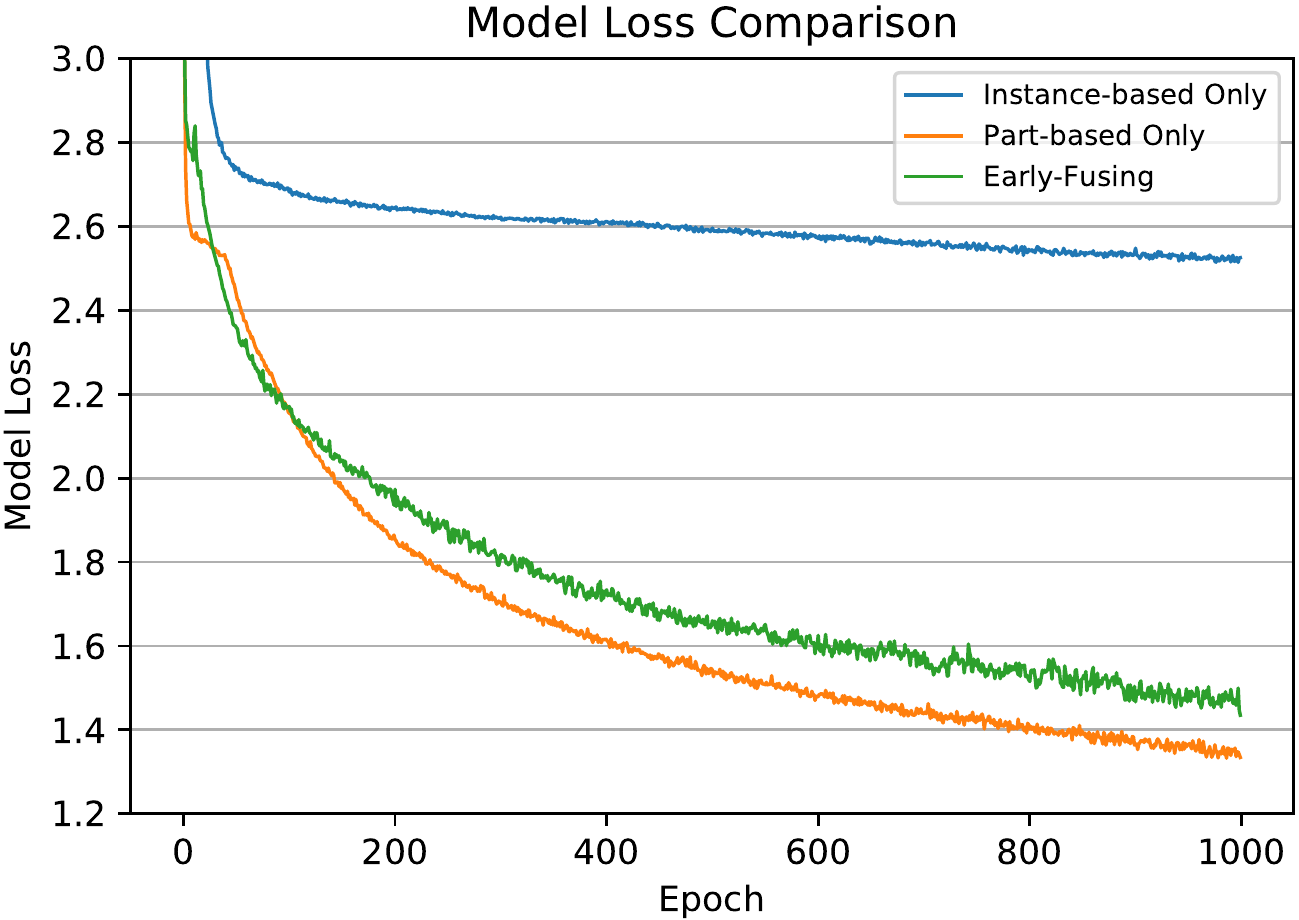}} 
   \subfigure[Test Accuracy]{ 
     \label{Figure:exp1_acc} 
     \includegraphics[width=0.45\textwidth]{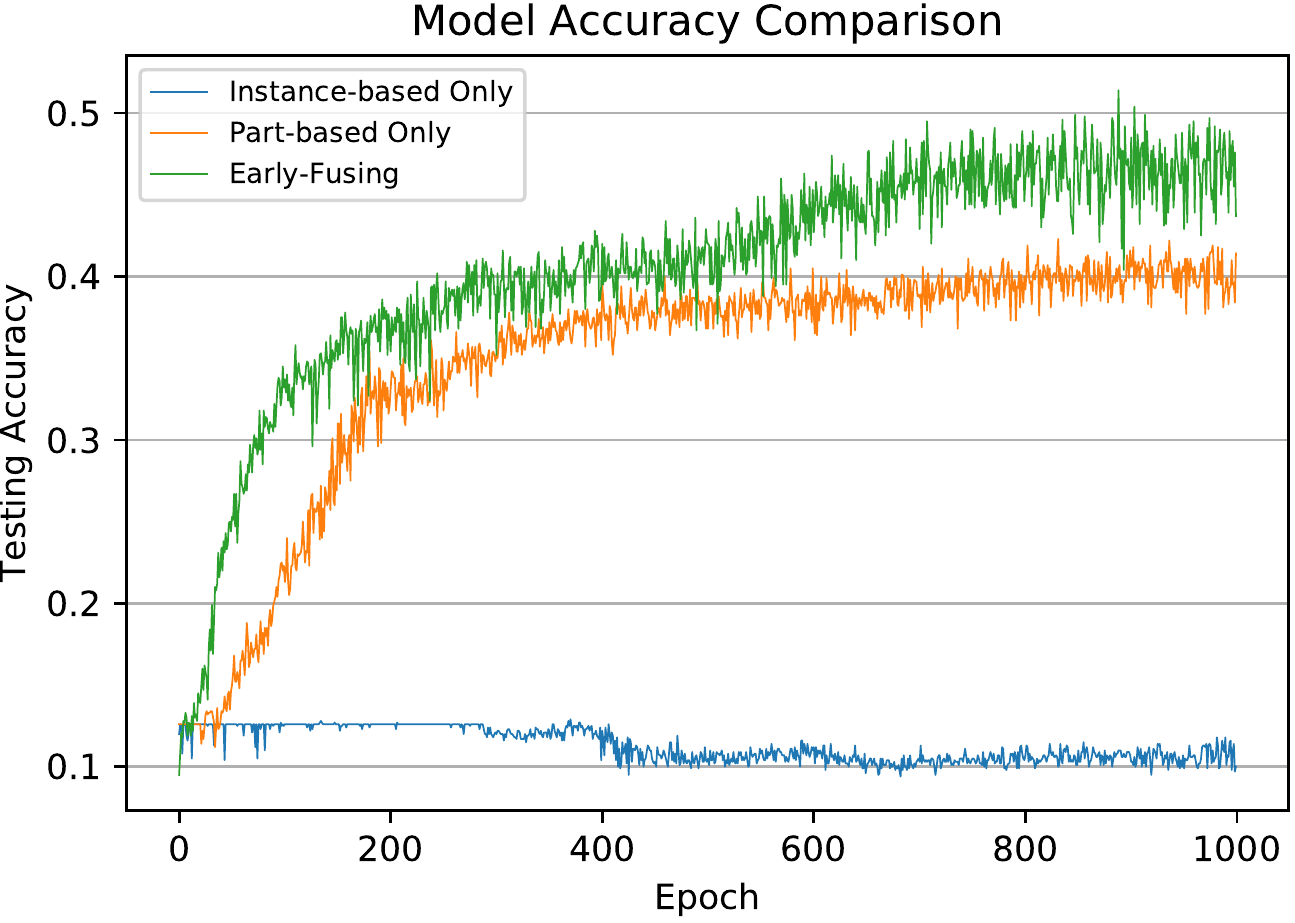}} 
   \caption{Comparison of loss and accuracy in MNIST-Action.} 
   \label{Figure:loss_acc}  
\end{figure*}

\begin{figure*}[!ht]
 	\begin{center}
 		\includegraphics[width=0.8\textwidth]{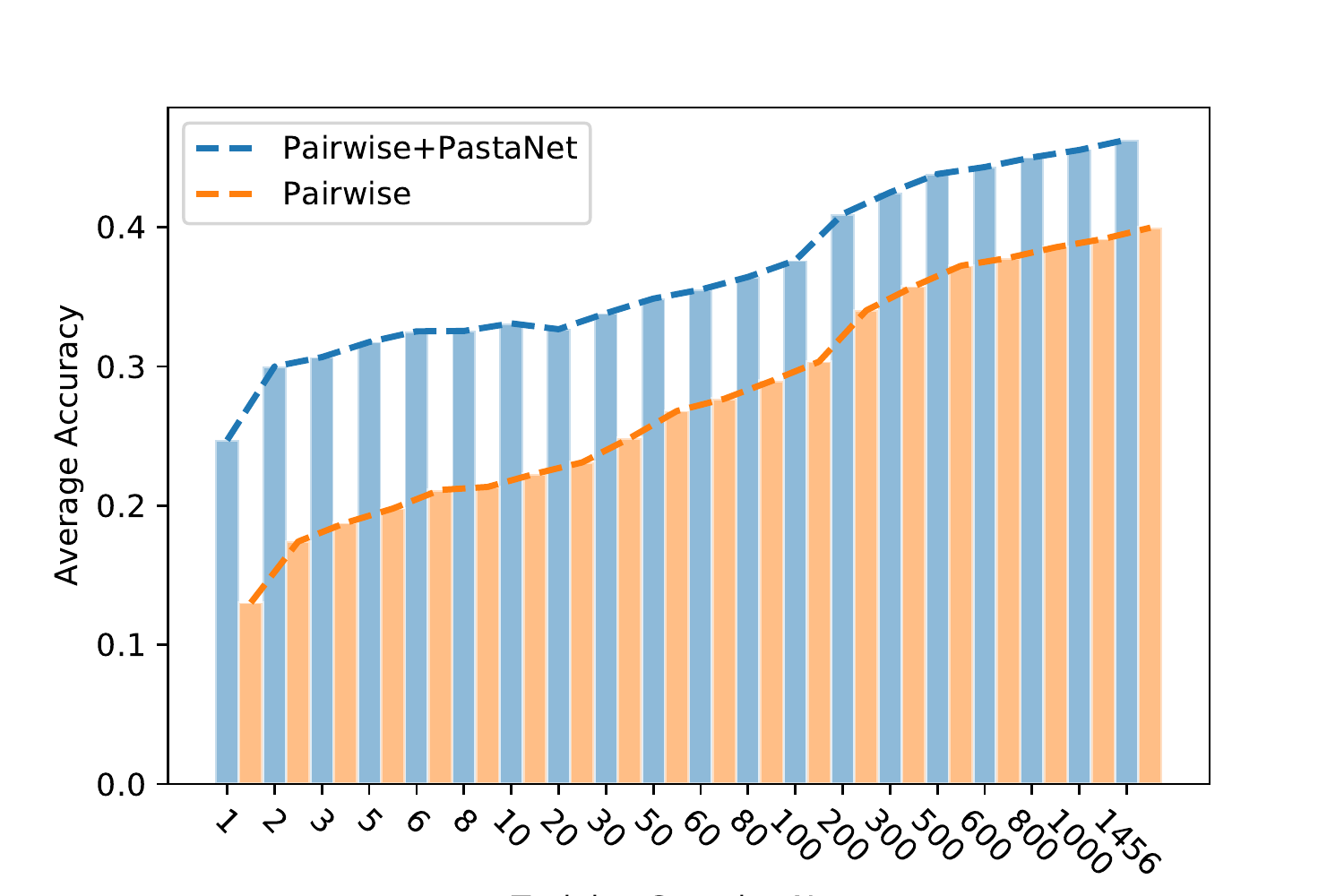}
 	\end{center}
 	\caption{Effectiveness on few-shot problems. The y-coordinate means the mAP of different sets. The x-coordinate indicates the activity sets. For example, $i$ means the number of training images is equal to or less than $i$, if $i$ is 1 then it means one-shot problem. Our approach can obviously improve the performance on few-shot problem, for the reason of reusability and transferability of \textit{PaSta}.}
 	\label{Figure:few}
\end{figure*}
Results are shown in Fig.~\ref{Figure:loss_acc}.
We can find that the hierarchical method largely outperforms the instance-based method.

\begin{figure*}[!ht]
 	\begin{center}
 		\includegraphics[width=0.9\textwidth]{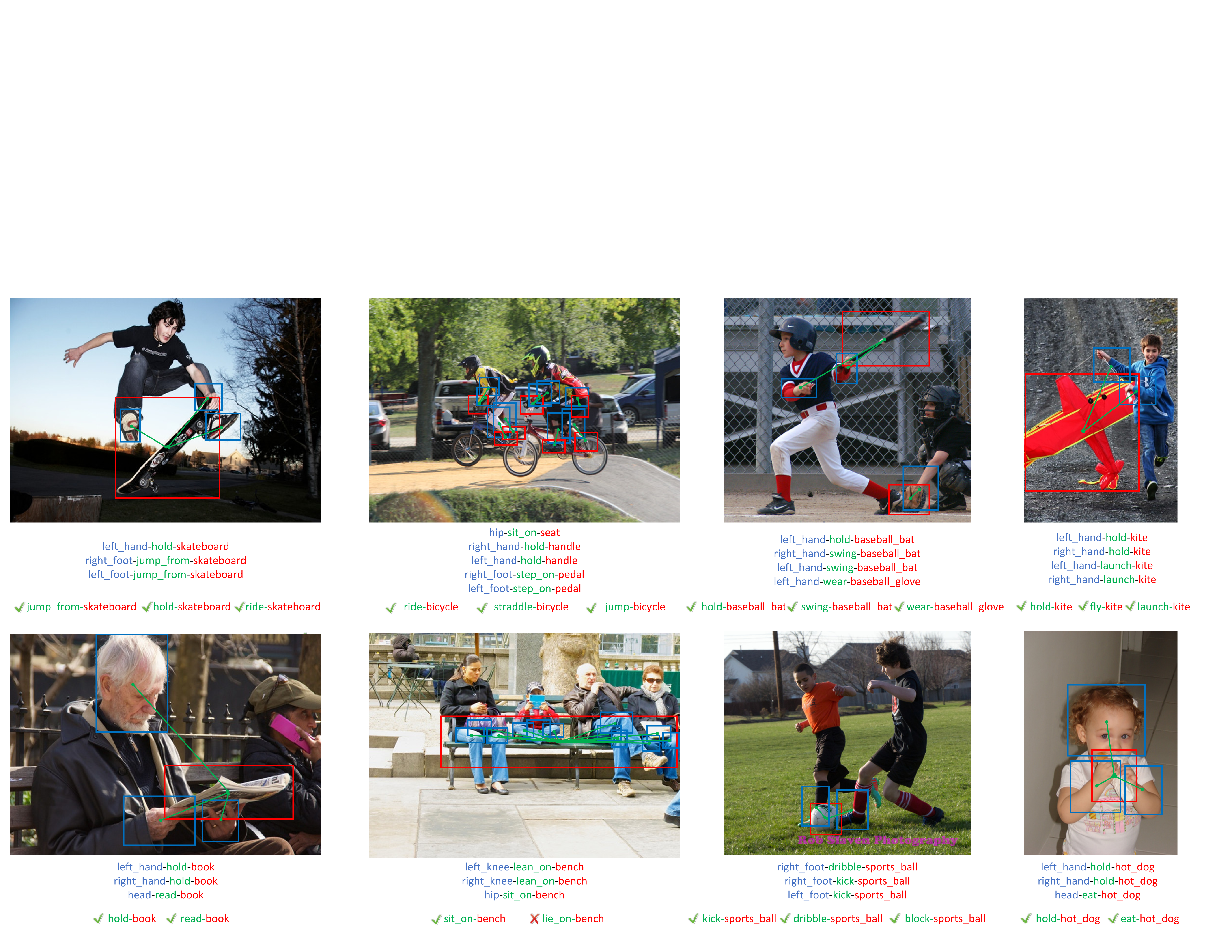}
 	\end{center}
 	\caption{Visualized results of our method. Triplets under images are predicted \textit{PaSta}. Human body part, verb and object are represented in blue, green and red. Green tick means right prediction and red cross is the opposite.}
 	\label{Figure:vis}
\end{figure*}

\section{Effectiveness on Few-shot Problems}
In this section, we show more detailed results on HICO~\cite{hico} to illustrate the effectiveness of \textit{Pasta} on few-shot problems.
We divide the 600 HOIs into different sets according to their training sample numbers.
On HICO~\cite{hico}, there is an obvious positive correlation between performance and the number of training samples. 
From Fig.~\ref{Figure:few} we can find that our hierarchical method outperforms the previous state-of-the-art~\cite{Fang2018Pairwise} on all sets, especially on the few-shot sets.
 
\section{Additional Activity Detection Results}
We report visualized \textit{PaSta} and activity predictions of our method in Figure~\ref{Figure:vis}.
The $\langle body\_part, part\_verb, object\rangle$ with the highest scores are visualized in blue, green and red boxes, and their corresponding \textit{PaSta} descriptions are demonstrated under each image with colors consisted with boxes. The final activity predictions with the highest scores are also represented.
We can find that our model is capable to detect various kinds of activities covering interactions with various objects. 

\begin{table*}[htb]
	\begin{center}
    \resizebox{0.48\textwidth}{!}{
        \begin{tabular}{l|l}
        Experiment & Pre-training Data\\
        \hline
        \textit{PaSta}Net* (Tab. 1)    & HICO train set\\
        \textit{PaSta}Net (Tab. 1)     & \textit{PaSta}Net w/o HICO test set\\
        \textit{PaSta}Net* (Tab. 2)    & HICO-DET train set \\
        \textit{PaSta}Net (Tab. 2)     & \textit{PaSta}Net w/o HICO-DET test set \\
        Tab. 3 & \textit{PaSta}Net w/o VCOCO \\
        Tab. 4 (AVA) & \textit{PaSta}Net\\ 
        Sec. 5.4   & \textit{PaSta}Net w/o HICO-DET \\
        \hline
        \hline
        Experiment & Finetuning Data (\textcolor{red}{Activity Labels})\\
        \hline
        Tab. 1    & HICO train set\\
        Tab. 2    & HICO-DET train set \\
        Tab. 3    & VCOCO train set \\ 
        Tab. 4 (AVA)   & AVA train set\\ 
        Sec. 5.4  & HICO-DET train set\\
		\end{tabular}}
	\end{center}
    \caption{Data usage. Tab. 3, 4 and Sec. 5.4 are transfer learning.}
    \label{tab:a2v}
\end{table*}

\section{Data usage}
The data usages of pre-training and finetuning are clarified in Tab.~\ref{tab:a2v}. We have carefully excluded the testing data in all pre-training and finetuning to avoid data pollution.

\end{appendices}

\end{document}